\documentclass[fleqn,10pt]{wlscirep}
\usepackage[utf8]{inputenc}
\usepackage[T1]{fontenc}
\usepackage{tikz-cd}
\usepackage[most]{tcolorbox}
\usepackage{float}
\usepackage{algpseudocode}
\usepackage{algorithm}
\usepackage{hyperref}
\usepackage{siunitx}
\usepackage{adjustbox}
\usepackage{cleveref}
\usepackage{graphicx}
\usepackage{svg}
\usepackage{shellesc}
\usepackage{array}
\usepackage{tabularx}
\usepackage{wrapfig}
\usepackage{enumitem}
\usepackage{listingsutf8}
\usepackage{xcolor}
\usepackage{placeins}
\usepackage{array}
\usepackage{subcaption}

\lstdefinelanguage{json}{
    basicstyle=\ttfamily\footnotesize,
    numbers=none,
    numberstyle=\tiny,
    stepnumber=1,
    numbersep=5pt,
    breaklines=true,
    showstringspaces=false,
    frame=single
}
\usepackage[backend=biber]{biblatex}
\usepackage{makecell}
\tcbset{
  colback=gray!5,
  colframe=black!60,
  fonttitle=\bfseries,
  coltitle=black,
  boxrule=0.8pt,
  arc=2pt,
  outer arc=2pt,
  left=6pt,
  right=6pt,
  top=6pt,
  bottom=6pt,
}

\lstdefinestyle{codeblock}{
    basicstyle=\ttfamily\footnotesize,
    numbers=none,
    numberstyle=\tiny\color{gray},
    stepnumber=1,
    numbersep=6pt,
    frame=single,
    rulecolor=\color{black!50},
    breaklines=true,
    breakatwhitespace=false,
    columns=fullflexible,
    keepspaces=true,
    showstringspaces=false,
    tabsize=4,
    captionpos=b
}

\lstdefinestyle{prompt-style}{
    backgroundcolor=\color{gray!10},
    basicstyle=\ttfamily\small,
    breaklines=true,
    frame=single,
    captionpos=b,
    numbers=none,
    inputencoding=utf8,
    extendedchars=true,
}

\lstdefinestyle{pythonblock}{
    style=codeblock,
    language=Python
}

\lstdefinestyle{jsonblock}{
    style=codeblock,
    language=json
}

\newtcolorbox{promptbox}[2][]{%
  title=#2, #1}

\newtcolorbox{optimizerbox}{
    enhanced,
    breakable,
    colback=white,
    colframe=green!60!black,
    colbacktitle=green!75!black,
    coltitle=white,
    fonttitle=\bfseries,
    title=Optimization Agent,
    boxrule=0.8pt,
    arc=2pt,
    left=3pt,
    right=3pt,
    top=3pt,
    bottom=3pt
}

\newtcolorbox{cypherbox}{
    enhanced,
    breakable,
    colback=white,
    colframe=blue!60!black,
    colbacktitle=blue!60!black,
    coltitle=white,
    fonttitle=\bfseries,
    title=Cypher Validation Agent,
    boxrule=0.8pt,
    arc=2pt,
    left=3pt,
    right=3pt,
    top=3pt,
    bottom=3pt
}

\newtcolorbox{schemebox}{
    enhanced,
    breakable,
    colback=white,
    colframe=orange!80!black,
    colbacktitle=orange!80!black,
    coltitle=white,
    fonttitle=\bfseries,
    title=Scheme Validation Agent,
    boxrule=0.8pt,
    arc=2pt,
    left=3pt,
    right=3pt,
    top=3pt,
    bottom=3pt
}

\title{LLMs in Process Diagram Engineering: From Optimal PFDs to Validated P\&IDs}

\author[1]{Timur Zakarin}
\author[1]{Sergei Voitov}
\author[1,*]{Sergei Shumilin}
\author[1,2]{Evgeny Burnaev}
\affil[1]{Skoltech, AI Center, Moscow, Russia}
\affil[2]{AIRI, Moscow, Russia}

\affil[*]{s.shumilin@skoltech.ru}

\begin{abstract}
Nowadays, the creation of a process flow diagram (PFD) and its subsequent transformation into a piping and instrumentation diagram (P\&ID) is predominantly performed manually. Applying artificial intelligence (AI) in the task could potentially lead not only to process automation and time savings, but also to financial gains by exploring numerous diagram's topology options and reducing manual labor. This research presents P\&ID Pilot - a practical end-to-end AI pipeline capable of handling flowsheet developing for both stages. The first stage focuses on PFD synthesis, whereas the second is directed toward modifying the generated PFD into P\&ID. After comparing four different methods, the hybrid approach combining genetic algorithms (GA) and large language models (LLM) is shown to generate the optimal valid PFD topology, achieving the lowest loss value among all the methods, while satisfying the required outlet flow parameters without engineering-rule violations. For the second stage, the proposed LLM-based agent successfully transforms the generated PFD into a source-grounded P\&ID by producing validated, executable modifications through a restricted engineering software development kit (SDK), achieving 100\% execution success while maintaining compliance with domain-specific rules and reference graph structures. This unified pipeline - coupling GA/LLM-driven synthesis with an LLM-based transformation agent - offers a feasible path toward end-to-end process design automation by producing validated, deployable outputs and substantially reduces manual engineering effort.


\end{abstract}
\begin{document}

\flushbottom
\maketitle
%
%
\thispagestyle{empty}

\section{Introduction}

Process flow diagrams (PFDs) and piping and instrumentation diagrams (P\&IDs) are central artefacts in process engineering. PFDs describe the main process topology, major equipment, and material streams, whereas P\&IDs provide a more detailed representation of process equipment, valves, piping, control structure, and instrumentation \cite{Lucia2008}. Together, these diagrams support design coordination, process analysis, safety review, documentation, and subsequent operation of industrial facilities. The need for consistent digital representations of such engineering information has motivated standardization efforts, including DEXPI-based data exchange for P\&IDs \cite{DEXPI2026} and graph- or sequence-based representations of process flowsheets such as SFILES 2.0 \cite{Vogel_2023}.

Despite the availability of computer-aided engineering tools, the development and modification of PFDs and P\&IDs remain a tedious and time-consuming task that offers significant potential for cost reduction and faster development cycles \cite{Uzuner2012}. Engineers must preserve process logic, equipment configuration constraints, connectivity, control structure, and diagram consistency while exploring alternative designs or applying local modifications. These operations are often repetitive but cannot be treated as purely mechanical, because even small changes in topology, attributes, or control links may invalidate downstream engineering assumptions. As a result, diagram preparation and revision are associated with considerable manual effort and cognitive load.

Recent work has begun to investigate artificial intelligence (AI) methods for related tasks. Data-driven models have been used for flowsheet autocompletion and control-structure prediction from process topologies \cite{vogel_2023,https://doi.org/10.1002/aic.18259,balhorn2024graphtosfilescontrolstructureprediction}. Reinforcement learning (RL) combined with graph convolutional neural networks (GCNN) has also been proposed for flowsheet synthesis and optimization \cite{Stops_2022, https://doi.org/10.1002/aic.18584}. In parallel, large language models (LLMs) have been applied to P\&ID understanding and generation, including natural-language interaction with DEXPI-derived graph representations \cite{Alimin_2025} and agentic creation of P\&ID diagrams from natural-language descriptions \cite{gowaikar2024agenticapproachautomaticcreation}. These studies indicate that AI can support process diagram engineering, but they also highlight the importance of structured representations and controlled interaction mechanisms.

The present work addresses the problem from a full-cycle perspective. We consider a technological diagram as a structured graph object whose nodes, edges, and attributes encode equipment, streams, connections, and engineering metadata. This representation provides a common computational substrate for optimization, model-based interpretation, and controlled modification. In contrast to unconstrained generation or direct editing of a serialized engineering file, the proposed P\&ID stage uses a software development kit (SDK)-bounded interaction model: an LLM interprets a user request and generates executable actions, while all access to the diagram is mediated by predefined graph operations and validation procedures.

\begin{figure}
    \centering
    \includegraphics[width=1.\linewidth]{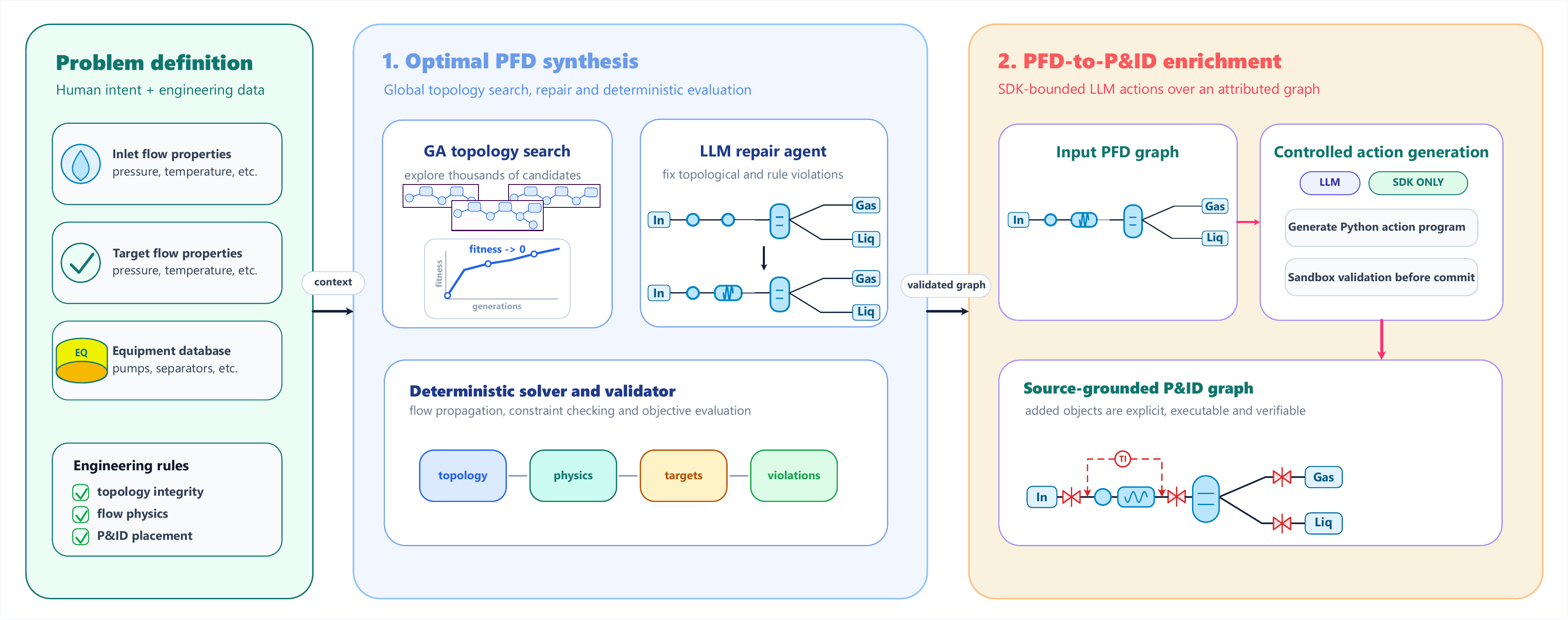}
    \caption{Visual abstract of the Full Cycle AI pipeline. \textbf{Core design principle}: The genetic algorithm combined with LLM repair for PFD synthesis + restricted-SDK P\&ID transformation enables full cycle process pipeline.}
    \label{fig:full cycle image}
\end{figure}

The contribution of this work is threefold. First, we present a pipeline that connects optimal PFD synthesis with subsequent P\&ID-oriented interaction. Second, we compare several approaches to PFD generation within a graph-based process-design setting. Third, we introduce a controlled LLM-based workflow for P\&ID analysis and modification, and evaluate it on domain-grounded executable rules over a graph representation of a P\&ID. The overall workflow is illustrated in Figure~\ref{fig:full cycle image}.

\section{Background}
There are several studies discussing the application of methods based either on the principles of AI or on basic mathematical principles for PFD and P\&ID analysis, generation, completion and editing.

\subsection{LLM usage}

\cite{Alimin_2025} discusses the ability of Large Language Model (LLM) to understand and analyze DEXPI XML P\&Ds. The authors suggest transforming DEXPI XML file into a graph representation that can be stored in Neo4j database as a knowledge graph. Afterwards, this knowledge graph is attached as GraphML file to the context when prompting an LLM with a question. Experiments prove that the LLM can accurately answer P\&ID-related questions.

The work \cite{gowaikar2024agenticapproachautomaticcreation} also primarily focuses on the DEXPI XML format. A specific Automatic Creation of P\&ID (ACPID) Copilot is developed that sequentially generates DEXPI XML based on a user's text description and combines several techniques such as a multi-step agentic workflow over a pre-trained LLM, domain specific language translation, human-in-the-loop validation, etc. The approach is compared with zero-shot and few-shot methods for the GPT-4-Turbo LLM model on soundness and completeness metrics and the authors conclude that the proposed Copilot achieves the best result on both metrics.

The next group of papers \cite{balhorn2024graphtosfilescontrolstructureprediction} and \cite{https://doi.org/10.1002/aic.18259} present a data-driven method for the prediction of control structures for flowsheets - the first step towards turning a PFD into a P\&ID. Though the goal of both papers is similar, they differ slightly in their architectural setup - specifically, input data handling. In \cite{balhorn2024graphtosfilescontrolstructureprediction}, a graph-to-sequence approach is presented: a Graph-to-SFILES model consisting of Graph Neural Network (GNN) as an encoder and a Transformer architecture as a decoder is applied. It takes the flowsheet topology as a graph input and returns a control-extended flowsheet as a sequence in the Simplified Flowsheet Input-Line Entry-System (SFILES) 2.0 notation \cite{Vogel_2023} (the format was adapted from the text-based Simplified Molecule Input-Line Entry-System (SMILES) - an example of natural language processing applications in chemistry \cite{https://doi.org/10.1002/aic.690310302}). In contrast, in \cite{https://doi.org/10.1002/aic.18259}, the authors study a sequence-to-sequence approach: the input topology of PFDs is transformed into strings using SFILES 2.0 notation (the vice-versa procedure is also used on the output to obtain graph data from the predicted strings) and the core architecture of the model is a T5 transformer which performs translation task in which PFDs without control structures are translated to PFDs with control structures. Both papers use quite the same principle in training and model evaluation to perform a fair comparison between the methods. Overall, both show quite promising results in which the graph-to-sequence model outperforms sequence-to-sequence model on the small datasets, whereas for larger datasets the opposite trend is observed.

A quite similar sequence-to-sequence approach for PFDs\ autocompletion is described in \cite{vogel_2023}. Inspired by the text translation task, the authors use a transformer-based language model that has only an auto-regressive decoder, while representing flowsheets in SFILES 2.0 notation. The model is pre-trained on the synthetic data and then fine-tuned via transfer learning step on real PFD topologies. The model's performance evaluations prove the concept's validity and suggest potential application for AI-aided assistance for chemical engineers. 

\subsection{RL usage}
\label{subsec: rl}

Another innovative approach for PFD creation is discussed in \cite{Stops_2022}. The authors study a combination of GCNN and RL for the optimal flowsheet synthesis task. Here, the GCNN is applied to flowsheets (which are used as graphs in this case) to represent them as numerical embeddings that are passed as input to the RL process. The latter consists of an actor-critic pair, an environment and a reward function based on ab economic formula. The core architectural components are Multi-Layer perceptron (MLP) and a GCNN. The results show that this method allows the generation of valid, economically optimal PFD. The same approach is also studied in \cite{https://doi.org/10.1002/aic.18584}. The main fundamental changes from the original paper are the use of novel proximal policy optimization and integration of masked agents. Tests on the creation of viable chemical process flowsheets demonstrate the efficiency of the proposed method.

\subsection{Genetic Algorithms}

Genetic algorithms (GA), first introduced by J. Holland in \cite{Holland:1975}, are actively used as optimization methods. Described as a stochastic methods and based on natural selection, these algorithms use an iterative search strategy to find an optimal solution via specific genetic operations (crossover, mutation, selection, etc.). In many works \cite{ahmadi2016performance, MOURA20101461, 09df465a51d34520aea2b115ec357733}, the efficiency of GAs has been proven in nonlinear optimization tasks with non-trivial objective function and various constraints with no exception for optimal PFD creation \cite{inproceedings}. Here GA is applied to generate the optimum configuration of a hybrid membrane/cryogenic system for a three membrane unit process. The results demonstrate the potential of the technique.

\section{Optimal PFD Synthesis}

Process flowsheet synthesis is the systematic application of computational techniques to generate process flowsheets by optimizing design and operation variables, equipment types, and process configuration \cite{https://doi.org/10.1002/aic.690310302}.

In this section we perform comparison of four various methods for PFD synthesis - a multi-agent LLM approach inspired by \cite{Alimin_2025}, RL/GCNN-based approach from \cite{Stops_2022}, GA  and hybrid GA/LLM approach - applied to a common test case to select the most suitable method based on accuracy, and optimization time.

The remainder of the section structured as follows: \ref{subsec:test case} provides a test case description, the following subsection \ref{subsec:solver} presents our custom Python solver, the subsequent subsections \ref{subsec:mas}, \ref{subsec:ga}, \ref{subsec:rl} and \ref{subsec:hybrid} describe each approach in detail, and \cref{subsec:results} presents a comparison and discussion of the received results.

\subsection{Test case description}
\label{subsec:test case}

To test the functionality of the system, a suitable task is selected that is moderately simple yet close to real-life processes. As a result the following scenario is proposed - generating the optimal PFD for an oil treatment unit (OTU). Such units are used for the  preliminary separation of crude oil from oil fields into oil, gas and formation water  to achieve  saleable quality.

The main operations in an OTU include gas separation, water separation, pressure increase, and flow heating.

The input data consist of an inlet flow properties (physical and chemical properties), a desired outlet flow properties specified by the user, and an equipment database containing various components used for PFD synthesis such as pumps, heat exchangers, water and gas separators, ball valves and pipe tees.

\subsection{Solver description}
\label{subsec:solver}

The solver is implemented as a Python module for calculating flows and checking constraints in the generated PFD. After constructing the graph, the solver propagates flow properties through all nodes of the graph.

Separate Python functions are implemented for each type of equipment, such as calculating of pressure increase in pumps according to specified Q-H characteristics, calculating of temperature changes in the heat exchanger based on a simplified heat balance, etc. More detailed information on the equations used in the calculation process is provided in Appendix \ref{app:equations} Additionally, the solver supports branching and combining flows. The detected violations of constraints are stored in the nodes of the graph and then taken into account in the loss (fitness) function. A general algorithmic description of the solver is provided in Appendix \ref{app:algorithms} in addition with the loss calculation function.

\subsection{Multi-agent system approach}
\label{subsec:mas}

It should be noted that the architecture in \cite{Alimin_2025} allows the LLM only to see and understand the entire flowsheet, but not to make changes according to user requests. However, this can be implemented by sending specialized Cypher requests.
Cypher is a declarative query language for graphs that allows to perform efficient queries on data in a property graph \cite{neo4jIntroductionCypher}. Given that LLM sees and understands the entire flowsheet as part of the context, it can be assumed that it will be able to make changes to the graph representation by sending generated Cypher requests. 

\begin{figure}[H]
    \centering
    \includegraphics[width=0.80\linewidth]{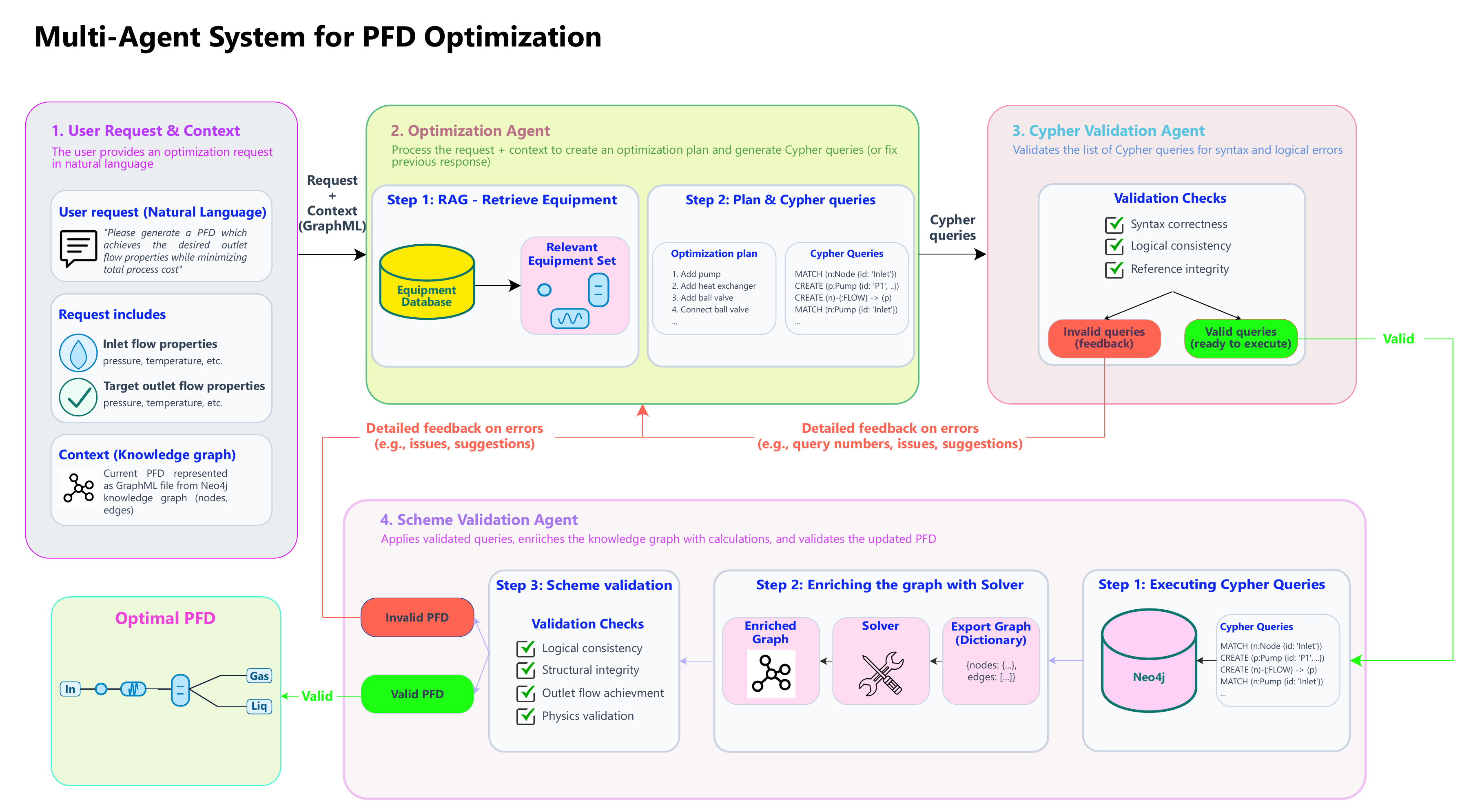}
    \caption{Architecture of the proposed MAS. \textbf{The Optimization Agent} receives the user's natural-language request along with the knowledge graph as additional context and processes it in two steps: retrieves relevant equipment using RAG and generates an optimization plan involving using the retrieved equipment and a list of corresponding Cypher queries. \textbf{The Cypher Validation Agent} checks the generated queries for any errors and, if errors are detected, generates feedback and sends it back to the \textbf{Optimization Agent}; otherwise, the validated query list is passed to the next agent. \textbf{The Scheme Validation Agent} executes the queries and checks the updated knowledge graph (i.e., the generated PFD) for compliance with engineering and topology constraints, If violations are detected, the feedback is passed to the \textbf{Optimization Agent} to correct the detected violations. If all checks are passed, the valid scheme is returned to the user.} 
    \label{fig:mas architecture}
\end{figure}

Many publications \cite{FRAGA2025109258, Kaven2024, yin2025} confirm the effectiveness of using not a single LLM, but a group of LLM agents where each agent performs a specific task while interacting with other agents across various engineering problems, with no exception of the chemical industry \cite{rupprecht2025multiagentsystemschemicalengineering}. Such a system is called Multi-agent system (MAS).
Thus, in our case, the following MAS architecture will be used, as shown in Figure \ref{fig:mas architecture}.

The whole system works in the following way. A user sends an optimization request which should contain initial flow properties and target flow properties in natural language to the system. The request is merged with a context that, in this case, is a GraphML file - a description of the current graph from Neo4j. It is then processed by an Optimization Agent in two steps. Firstly, it uses the Retrieval Augmented Generation (RAG) technique to extract relevant equipment from an equipment database which is preliminarily populated by the user. Secondly, given the retrieved equipment set, the agent composes an optimization plan and a list of Cypher queries to build the PFD according to the plan.
Then this list is passed to the next agent - the Cypher Validation Agent to assess the queries for syntax and logic errors. If there are any errors, it gives detailed feedback on them to the Optimizer so the queries could be fixed and passed again. The Cypher validation phase is followed by the scheme validation step which consists of three parts: execution of Cypher queries, enriching the updated knowledge graph with the solver, and a PFD check by the Scheme Validation Agent. The list of validated Cypher queries is applied, thus the knowledge graph stored in Neo4j is turned into the planned PFD. Afterwards, this new diagram is exported as a graph and passed as a dictionary of nodes, edges and their attributes to the solver. The solver performs flow calculations and produces an enriched graph and the latter is then analyzed by the Scheme Validation Agent for the presence of orphaned nodes, structural integrity, reaching and whether the target outlet flow properties are reached.

\begin{figure}[H]
    \centering
    \includegraphics[width=0.6\linewidth]{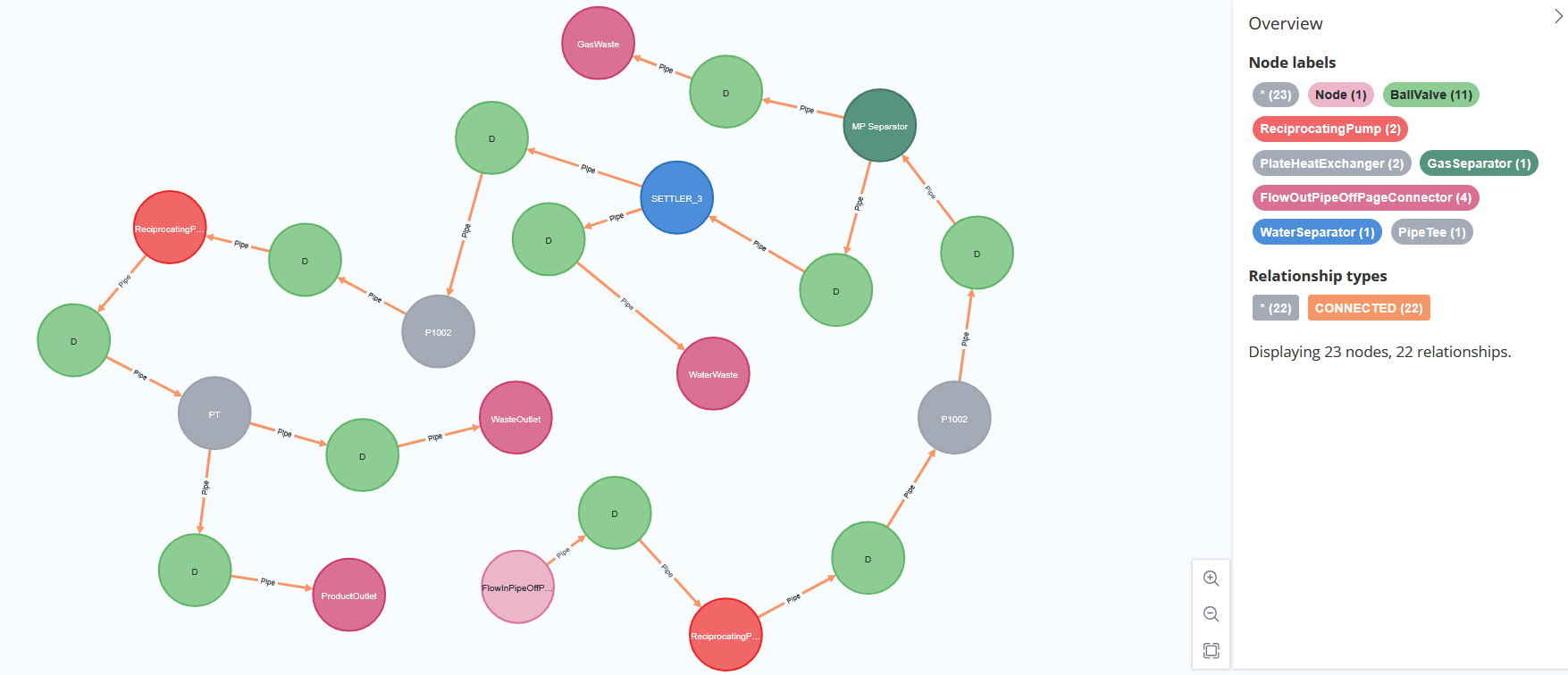}
    \includegraphics[width=0.9\linewidth]{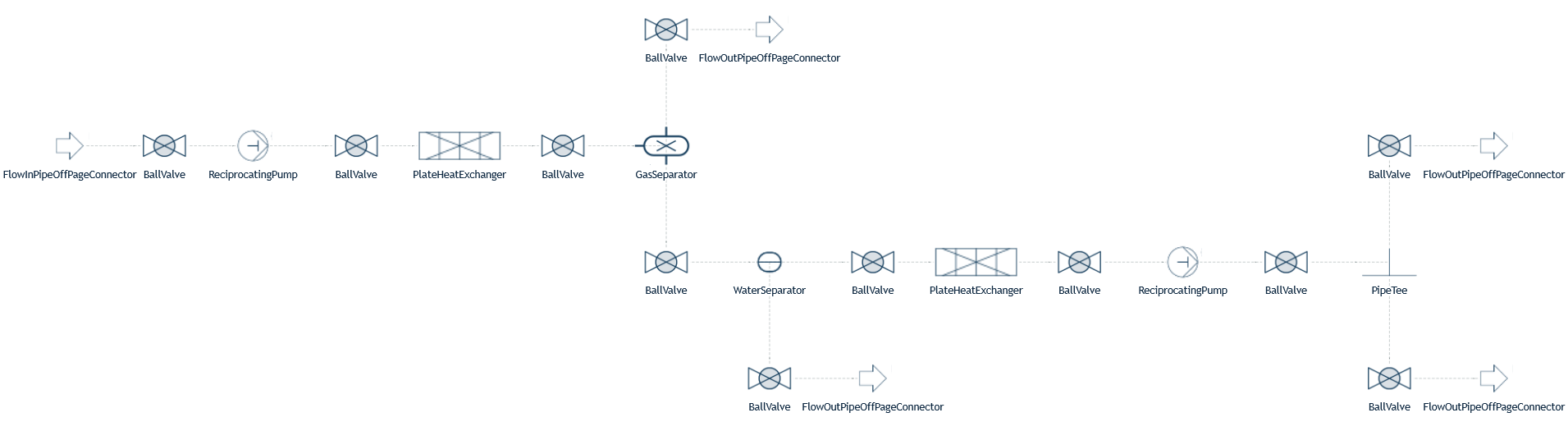}
    \caption{PFD of the OTU generated by the proposed MAS (model - \textbf{gpt-oss:120B}). \textbf{Top}: knowledge graph representation stored in the Neo4j graph database; \textbf{Bottom}: graph representation of the corresponding PFD. Overall, the generated diagram contains \textbf{23 nodes and 22 edges. No engineering or topology violations are detected}}
    \label{fig:pfd generated by mas}
\end{figure}

The suggested system was tested not only on the test case but on various LLMs with different parameter sizes as well. The models are served locally via vLLM.  The following LLMs were tested: Qwen3.6-35B-A3B \cite{qwen36_35b_a3b}, gpt-oss:120B \cite{openai2025gptoss120bgptoss20bmodel}, Qwen3.5-397B-A17B-FP8 \cite{teamqwen3} and DeepSeek-V4-Pro \cite{paper82956}.
For a fair LLM comparison, all agents' system prompts remained unchanged  as well as the temperature parameters. A sample of agents' interaction is presented in Appendix \ref{app:agents interaction}. The comparison results are shown in Table \ref{tab:mas_llm_comparison}.
A sample of a generated PFD is depicted in Figure \ref{fig:pfd generated by mas}.

Overall, the best results are demonstrated by the Qwen3.6-35B-A3B and gpt-oss:120B models: despite the PFD from the Qwen model being more optimal, it has minor issues with closing its branches, while the diagram from the other model is closer to the valid scheme. On the contrary, the larger models struggled to converge and provide a valid generated PFD.

\begin{table}[H]
\centering
\caption{Comparison of various LLM for MAS.}
\label{tab:mas_llm_comparison}
\begin{adjustbox}{width=\textwidth}
\begin{tabular}{lllll}
\toprule
\textbf{Models} & \textbf{Qwen3.6-35B-A3B} & \textbf{gpt-oss:120B} & \textbf{Qwen3.5-397B-A17B-FP8} & \textbf{DeepSeek-V4-Pro} \\
\midrule
Parameter size, B & 35   & 120 & 397 & 1600 \\
Context size (knowledge graph size) & 2101 (1293) & 2101 (1293) & 2101 (1293) & 2101 (1293) \\
Number of queries to LLM & 13 & 35 & 13 & 13 \\
Number of iterations & 6 & 13 & 7 & 7 \\
Work time, sec & 284.930 & 875.607 & - & - \\
Loss value & 186342.936 & 212997.603 & - & - \\
Comments & Incomplete PFD & \textbf{Valid PFD}  & Could not converge & Could not converge \\
\bottomrule
\end{tabular}
\end{adjustbox}
\end{table}

\newpage

\subsection{Genetic algorithm}
\label{subsec:ga}

\begin{figure}[H]
    \centering
    \begin{minipage}[c]{0.40\textwidth}
A standard GA workflow consists of the following iterative steps \cite{Gad2023-hn}:
       \small
        \setlist[enumerate]{itemsep=4pt, parsep=0pt, topsep=0pt}
        \begin{enumerate}
            \item An initial population of random chromosomes (topologies) is generated.
            \item A fitness (loss) value is calculated for each chromosome.
            \item Chromosomes with the best fitness values are selected from the current population for further operations (the selection method is treated as a hyperparameter).
            \item A new set of chromosomes is created by applying genetic operations (crossover, mutation, etc.) to the selected chromosomes.
            \item The population is updated with the new set, and Steps 2--4 are repeated until the maximum number of generations is reached.
        \end{enumerate}
    \end{minipage}
    \hfill
    \begin{minipage}[c]{0.56\textwidth}
         \centering
        \includegraphics[width=0.8\textwidth]{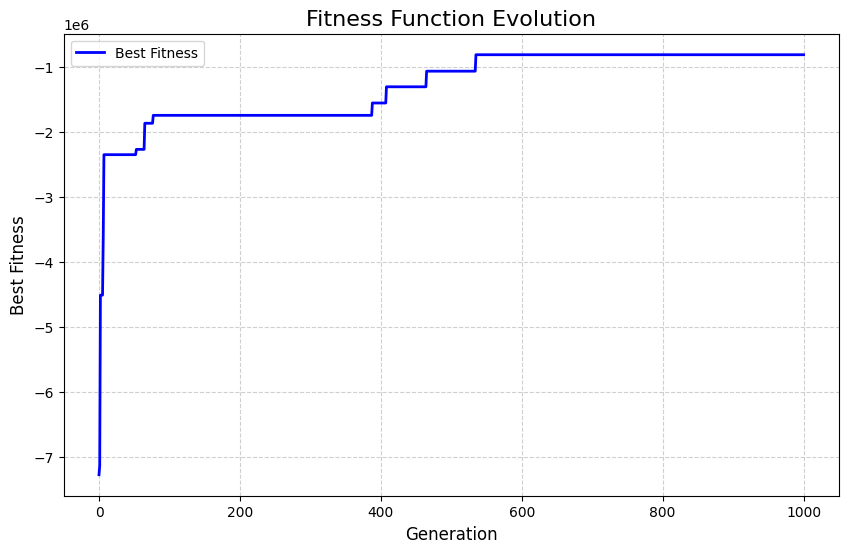}
        \captionsetup{font=small}
        \caption{Evolution of the GA fitness value during optimization. Over \textbf{1000 generations, corresponding to the evaluation of 780,800 chromosomes}, the fitness \textbf{is reduced by more than a factor of seven.} The convergence trend demonstrates the ability of the GA to progressively improve candidate PFDs.}
        \label{fig:ga_plot}
    \end{minipage}
\end{figure}

\begin{figure}[H]
    \centering
    \includegraphics[width=1.0\linewidth]{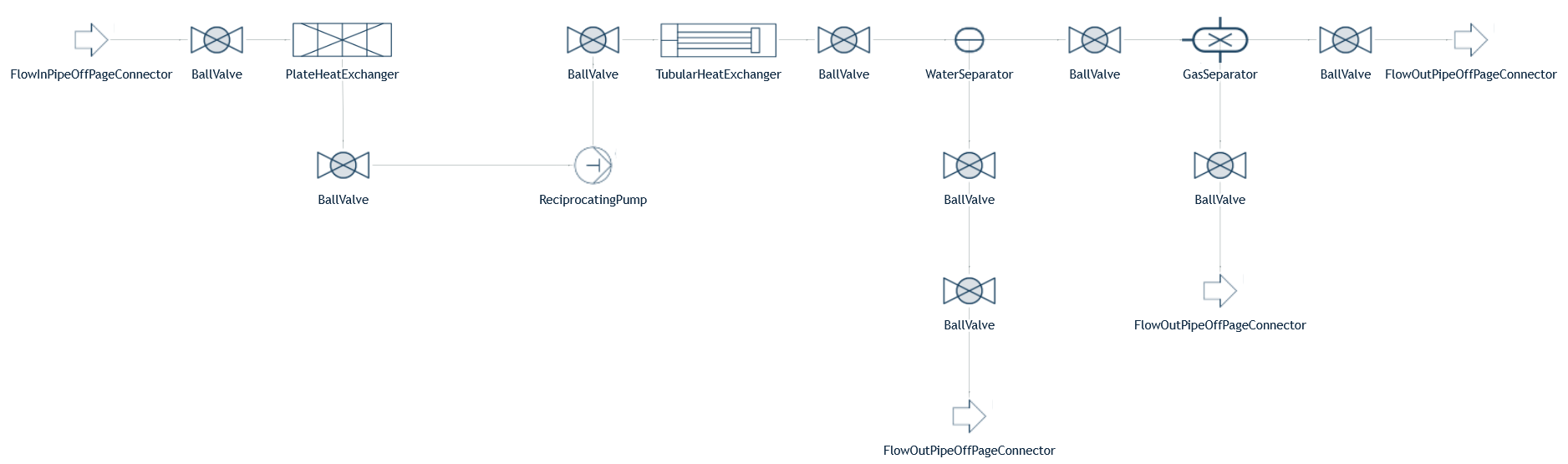}
    \caption{PFD of the OTU generated by the GA. Overall, the generated diagram contains \textbf{18 nodes and 17 edges.} Although the optimization significantly improves the objective value, node-level validation violations remain, indicating that the generated PFD does not fully satisfy the engineering constraints.}
    \label{fig:ga_graph}
\end{figure}

\noindent For correct fitness calculation, all evaluated chromosomes are converted into graph representation.
The experiments for GA evaluation are conducted using the specific Python package PyGAD \cite{Gad2023-hn}. The plot demonstrating the fitness function evolution through the number of generation is shown in Figure \ref{fig:ga_plot} and the PFD generated by the GA is depicted in Figure \ref{fig:ga_graph}.

\subsection{Reinforcement learning and Graph Convolutional Neural Networks}
\label{subsec:rl}

This approach is inspired by two key works: \cite{Stops_2022} and \cite{kwon2023rewarddesignlanguagemodels}. The first one introduces the whole concept of RL and GCNN application for flowsheet synthesis as discussed in Section \ref{subsec: rl} while the other paper studies the consept of LLM-as-a-reward in RL. We enhanced the original idea by introducing an LLM-as-a-reward policy mainly to increase system flexibility.

The whole system functions in the following way. In the beginning the developed graph is transformed into a numerical embedding - a flowsheet fingerprint. Afterwards the fingerprint is passed to the Actor, which attempts to extend the current graph. This decision is made on three levels: at the level 1, the Agent using a GCNN selects an open stream to replace with equipment; at level 2 the equipment type is chosen using an MLP (discrete decision); and finally, during the third level, the equipment design variables are defined (continuous decision). It is worth noting that the decision is made by an MLP, and a separate MLP has been developed for each type of equipment. The extended graph is then passed to the Critic to evaluate the Actor performance. This process continues iteratively. A more detailed description of every aspect can be found in the original paper \cite{Stops_2022}.

\begin{figure}[H]
    \centering
    \includegraphics[width=0.85\linewidth]{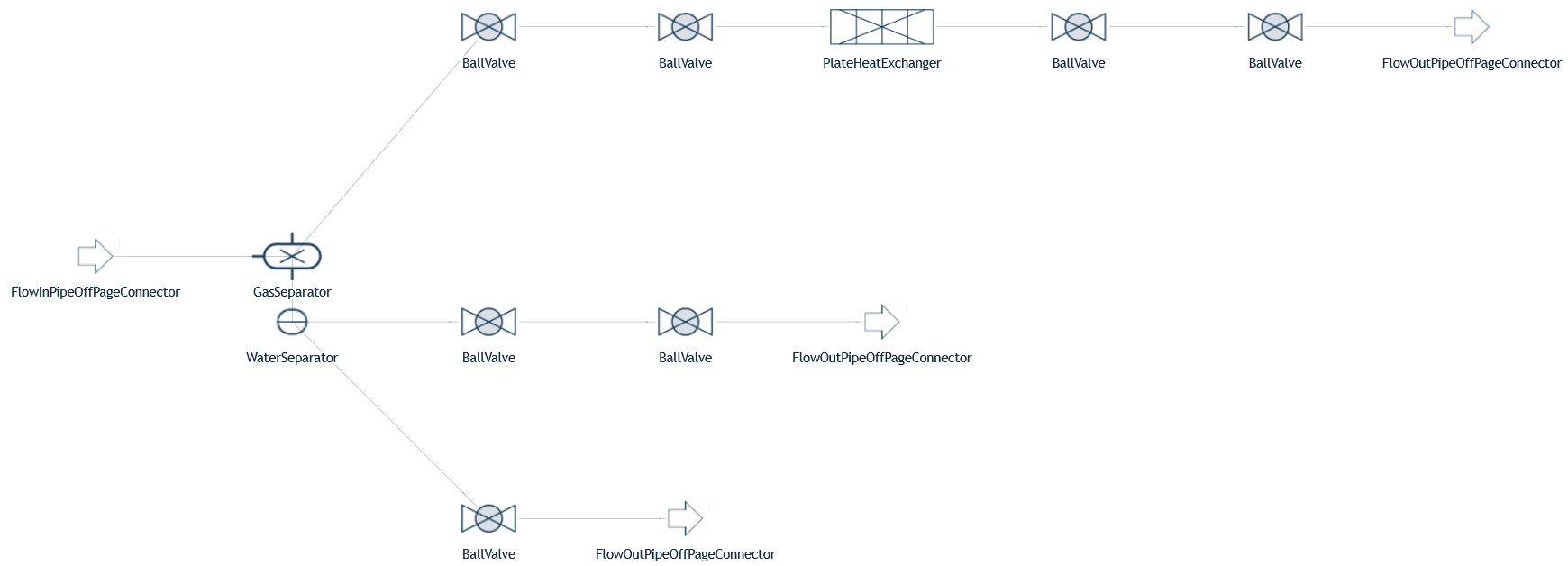}
    \caption{PFD of the OTU generated by the RL/GCNN approach. Overall, the generated diagram contains \textbf{14 nodes and 13 edges.} Although the resulting topology is executable, several topology validation violations remain, preventing the generated process flow diagram from satisfying all engineering requirements.}
    \label{fig:pfd generated by RL}
\end{figure}

The DeepSeek-V4-Pro LLM model was used in this experiment. The PFD generated by RL/GCNN can be found in the Figure \ref{fig:pfd generated by RL}.

\subsection{Hybrid approach}
\label{subsec:hybrid}

As can be seen from the results above - every tested method has its own advantages. The GA approach is capable of evaluating thousands of topology options, while the LLM-based MAS has been proven to be effective at modifying existing topologies. Given these circumstances, we decided to additionally test a hybrid approach which merges these two strengths by combining the techniques. Initially, an initial topology is generated by the GA. Afterwards, it is then modified by the LLM (the used model is gpt-oss:120, the prompt is provided in Appendix \ref{app:llm prompt hybrid method}) to fix any detected errors provided by the deterministic validator. 

\begin{figure}[H]
    \centering
    \includegraphics[width=0.95\linewidth]{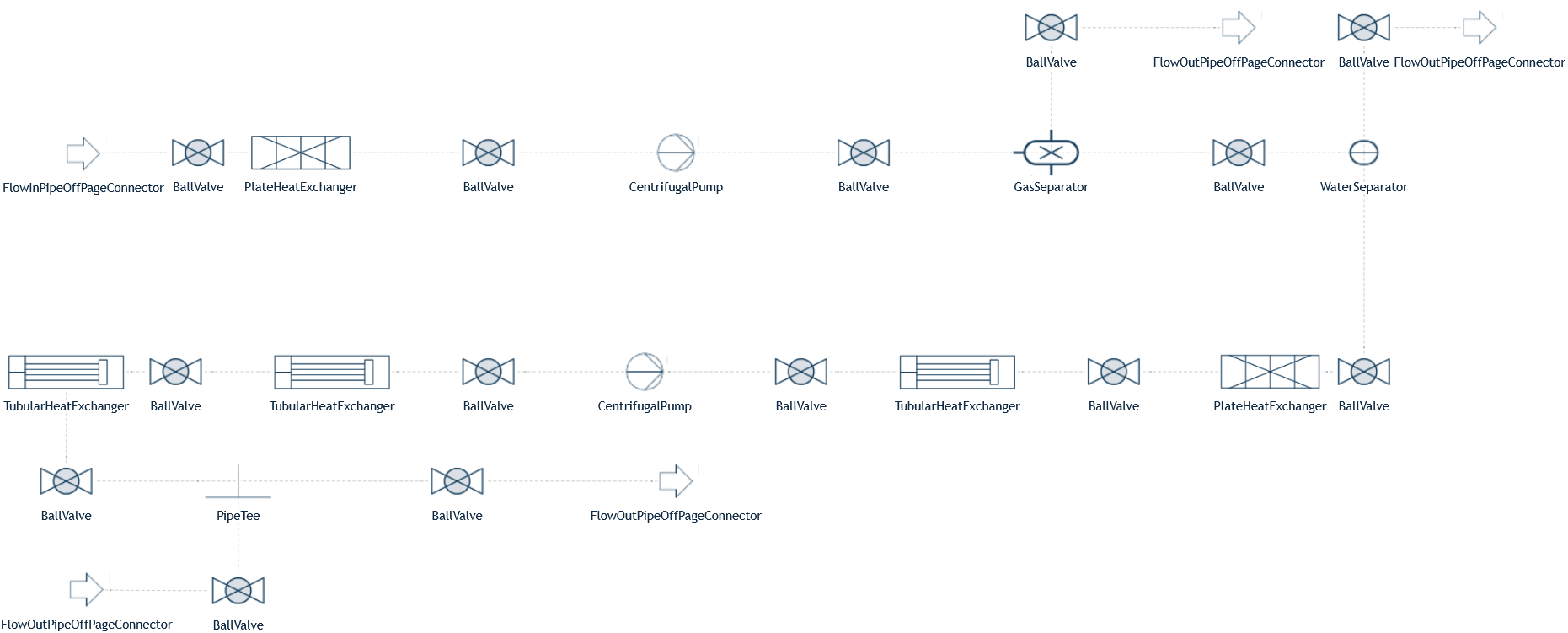}
    \caption{PFD of the OTU generated by the hybrid GA/LLM approach. Overall, the generated diagram contains \textbf{29 nodes and 28 edges.} No engineering or topology violations are detected, demonstrating that the proposed hybrid framework successfully generates a fully valid PFD.}
    \label{fig:pfd generated by hybrid}
\end{figure}

The final valid scheme received from the hybrid approach is shown in Figure \ref{fig:pfd generated by hybrid}.

\subsection{Results comparison}
\label{subsec:results}

The obtained results for all methods are shown in Table \ref{tab:comparison table pfd}. It can be seen that the valid PFD with no violations that comes closest to all the target flow properties was produced by the hybrid approach. The remaining methods generate the PFDs that either do not come as close to the target flow property values as the hybrid method to target flow properties values, or have violation  issues (topological and/or physical).

\begin{table}[H]
    \centering
    \caption{Comparison of optimal PFD generation approaches.}
    \begin{tabularx}{\textwidth}{llllX}
    \hline
         \textbf{Method} & \textbf{Time to generate, s} & \textbf{Loss Metric} & \textbf{Cost} & \textbf{Comments} \\ \hline
         GA (\textbf{Baseline}) & \textbf{505.6} & $0.808 \times 10^6$  & 31040 & Higher outlet deviation; has node violations \\
         MAS & 875.6 & $0.213 \times 10^6$ & 42900 & Higher outlet deviation \\
         RL/GCNN & 1862.2  & $55.939 \times 10^6$ & 17150 & Higher outlet deviation; has topology violations \\
         Hybrid (GA/LLM) & 2382.5 & $\mathbf{0.089 \times 10^6}$ & 61550 & \textbf{Lowest outlet deviation} \\ \hline
    \end{tabularx}
    \label{tab:comparison table pfd}
\end{table}

\section{P\&ID}

\subsection{LLM-Based Interaction with the P\&ID Model}

\subsubsection{Motivation and Relation to Existing Work}

The transition from a PFD to a practically useful P\&ID requires more than adding graphical symbols to an existing flowsheet. A P\&ID combines process topology, piping connectivity, equipment identifiers, valve and instrument placement, control links, and domain-specific design rules. Consequently, a language model cannot be treated as a free-form diagram editor: an incorrect object identifier, unsupported operation, inconsistent connection, or invalid attribute may compromise the consistency of the engineering model.

Recent work has shown that P\&ID information can be made accessible to LLMs by converting DEXPI representations into graph-based knowledge structures \cite{Alimin_2025}. Other approaches investigate the automatic creation of P\&IDs from natural-language descriptions through multi-step agentic workflows \cite{gowaikar2024agenticapproachautomaticcreation}, or predict control structures from PFD topologies using SFILES-based generative models \cite{https://doi.org/10.1002/aic.18259,balhorn2024graphtosfilescontrolstructureprediction}. These studies provide important evidence that generative models can operate on process-engineering diagrams. However, practical P\&ID interaction also requires a mechanism that prevents the model from modifying the diagram outside a controlled set of engineering operations.

The design principle adopted in this work is therefore to separate natural-language interpretation from diagram modification. The LLM is used to interpret a user request, identify relevant diagram objects, and generate an executable Python action program. The program can access the P\&ID only through a restricted software development kit SDK, while the actual graph operations are performed by deterministic procedures. This architecture preserves the flexibility of natural-language interaction, but constrains the model's action space and makes each proposed modification explicit, reproducible, and checkable before it is applied.

\subsubsection{Task Formulation}

The P\&ID interaction task is formulated as controlled action generation over a graph-based engineering model. The input consists of a current P\&ID graph, a natural-language user request, the public SDK contract, and optional domain hints or examples. The output is either an executable action program that analyzes or modifies the graph, or a structured result of a diagram-checking procedure.

The generated program must satisfy three constraints. First, it must use only documented SDK operations. Second, it must refer only to objects, attributes, and relation types available in the current diagram model or in the SDK contract. Third, it must be executable in an isolated validation environment before any modification is committed to the working diagram. Under this formulation, the LLM does not directly edit an image, XML file, CAD file, or serialized graph. It proposes a sequence of bounded operations whose validity can be checked by software.

\subsubsection{Graph Representation of the Diagram}

Internally, a P\&ID is represented as a directed attributed graph. Nodes correspond to diagram objects, including equipment units, pumps, valves, sensors, pipe elements, connection points, and auxiliary entities. Edges encode relations between these objects, such as physical connections, flow direction, control links, and other logical dependencies. This representation makes diagram topology available for algorithmic processing and allows local fragments of the diagram to be queried in a uniform way.

Nodes and edges may store engineering attributes when such information is available in the source representation. Examples include equipment tags, pipe classes, medium codes, flow directions, operating temperatures, pressures, and other process or documentation parameters. Thus, the graph model contains both connectivity information and part of the engineering context required for P\&ID analysis, rule checking, and controlled modification.

The operations considered in this study depend on the semantic and
topological structure of a diagram rather than on its original drawing
layout. We therefore use a layout-independent attributed graph in which
equipment, instruments, and valves are represented as nodes, while process
connections and attachment relations are represented as directed edges.
Drawing coordinates, symbol alignment, and graphical line routing are not
used during rule interpretation or validation.

For visualization, the graph is arranged automatically to make its
connectivity and local modifications easier to inspect. This representation
preserves the engineering information required by the evaluated rules,
which are expressed in terms of object types, attributes, adjacency,
directed order, paths, and attachment relations.

\subsubsection{SDK as a Layer of Controlled P\&ID Operations}

The SDK defines the boundary between LLM reasoning and executable diagram operations. In the current implementation, the agent interacts with the graph model through a predefined set of primitives for object search, attribute access, relation traversal, local topology extraction, and restricted graph modification. The model selects operations and parameters, but the semantics of these operations are defined by the SDK rather than by the language model.

The supported operations cover both analytical and editing scenarios. They include finding objects by type or attributes, reading node and edge properties, traversing incoming and outgoing connections, extracting local neighborhoods, adding objects and connections, inserting a new element into an existing connection, and attaching an element to an existing diagram object. More complex engineering actions, such as adding an instrument to equipment, inserting a valve into a line, or checking a local design rule, can be expressed as compositions of these primitives.

This decomposition is important for integration with CAD or engineering data systems. The high-level agent logic does not need to be rewritten for every target environment. Instead, the CAD-dependent layer is responsible for mapping SDK primitives to the corresponding external API calls, importing and exporting data, synchronizing changes, and preserving the internal graph representation. As a result, the LLM-facing contract can remain stable while implementation-specific data access is isolated in the lower layer.

Before execution, every generated action program is checked against the SDK contract. The program is then executed in an isolated environment on the current graph state. The resulting state or analytical output can be recorded, inspected, visualized, or used for subsequent validation. In this way, the SDK acts as both an operational interface and a safety boundary for LLM-based P\&ID interaction.

\subsubsection{Workflow of User, LLM, and Diagram Interaction}

The workflow starts with a natural-language request. The request may refer to diagram analysis, such as finding objects of a given type or checking a design rule, or to diagram modification, such as inserting a valve into a line or attaching a new instrument to an equipment item. The request is not applied to the diagram directly. It is first converted into an executable sequence of SDK calls.

At the interpretation stage, the LLM receives the current diagram context, the SDK contract, domain-specific hints, and examples of typical actions. Based on this information, it generates a Python action program that queries or modifies the P\&ID graph only through the available SDK operations. The generated program is then validated. The system checks whether only allowed operations are used and whether the program can be executed on the current graph without invalid object references, unsupported parameters, or runtime errors.

If validation fails, the diagram is left unchanged. The error information can be returned to the generation step, allowing the model to revise the program. If validation succeeds, analytical requests return a structured result, such as a list of found objects or rule violations. For modification requests, the validated action can be applied to the working graph after user confirmation, and the updated graph becomes the new diagram state.

The agent is therefore not treated as an autonomous P\&ID editor. Its role is to translate natural-language engineering requests into checkable actions over a formal diagram model. This preserves the usability of an LLM interface while limiting the model's influence through the SDK boundary, isolated execution, and validation procedure.

\begin{figure}[h]
    \centering
    \includegraphics[width=0.95\textwidth]{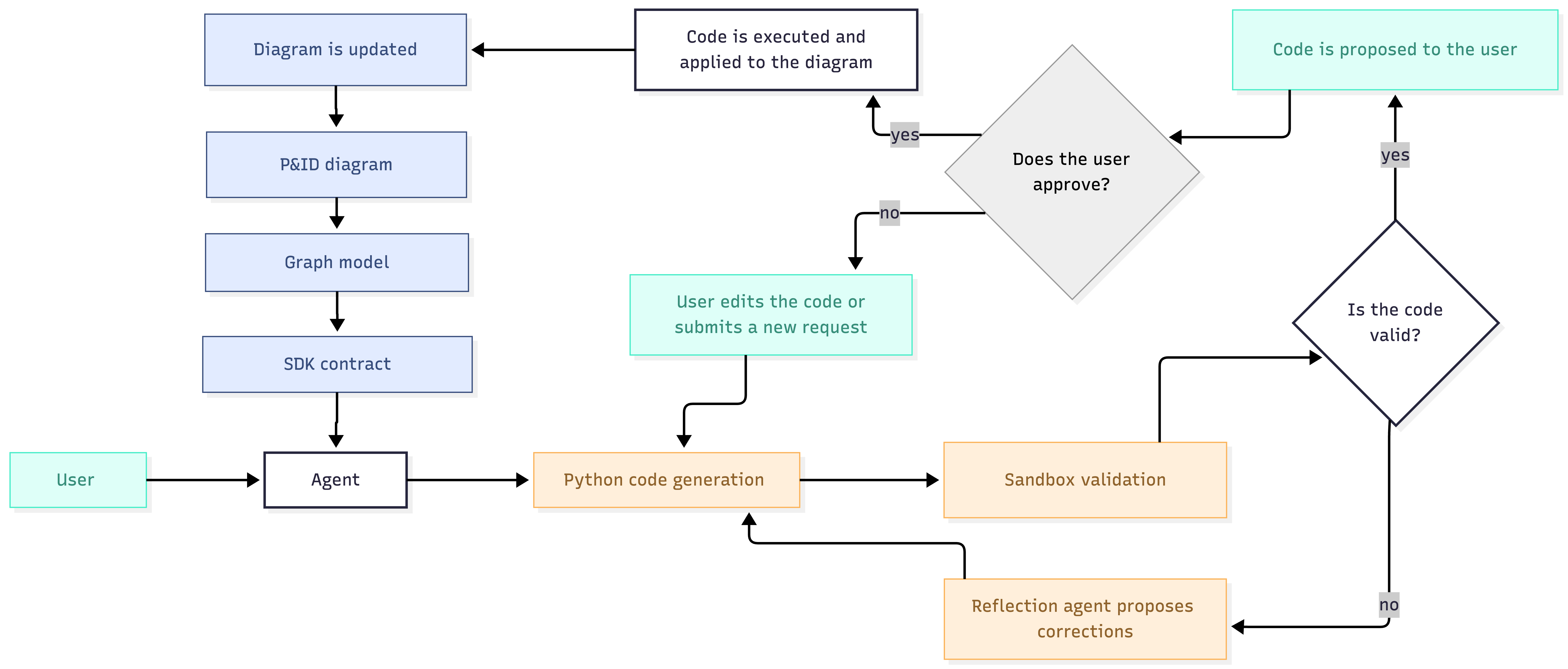}
    \caption{Workflow of user interaction with the P\&ID graph model through an LLM agent and a restricted SDK interface.}
    \label{fig:pid-llm-workflow}
\end{figure}

Figure~\ref{fig:pid-llm-workflow} shows the overall interaction between the user, the LLM agent, the SDK, the isolated validation environment, and the P\&ID graph representation. The user formulates an engineering task; the LLM constructs an action program using the available SDK interface; the system checks the correctness and executability of the actions; and, after user confirmation when required, the actions are executed on the graph model of the diagram.

\subsubsection{Checking and Controlled Diagram Modification}

LLM-generated actions are checked at two levels. The first level verifies SDK compliance: the program must use only documented operations, valid parameters, and object types available in the current graph model. This check is necessary because language models may generate plausible but unavailable calls or assume operations that are not implemented.

The second level evaluates executability and local effect. The generated actions are run in an isolated validation mode on the current diagram model. This run detects invalid object references, incorrect assumptions about graph structure, invalid traversal of connections, and modifications that cannot be applied. The main working version of the diagram is not changed during this stage.

Only after successful validation can a modification be applied to the working graph, and modification requests require user confirmation. Analytical requests, such as rule checks or object searches, return a structured result without changing the diagram state. This mechanism reduces the risk of uncontrolled edits while still allowing the LLM to support practical P\&ID analysis and controlled diagram modification.

\subsection{Experiments}
\label{sec:experiments}                                                 
                                                                      
\subsubsection{Domain-grounded validation}                                 
                                                                      
The evaluation is based on a set of engineering rules for checking P\&ID diagrams. These rules were formulated together with experts
in process design. We assess more than the quality of the generated     
code: we check whether the agent understands a rule given in natural    
language, maps it onto the graph model of a diagram, selects the        
relevant objects, and returns a result that can be verified             
automatically.                                                          
                                                              
Each scenario is executed as a runnable check over the graph. The       
agent receives the rule in natural language and obtains access to the
graph model through an SDK. It then generates code that performs the    
check and writes the result into a predefined variable. The code is     
executed in an isolated environment, and the returned value is          
compared with the reference answer stored in the test.                  
                                                                      
\subsubsection{Test set}
                                                                      
The model used in the experiments is a manually prepared diagram. It
covers typical fragments of process systems, including plant
boundaries, pump and compressor sections, column equipment, recycle     
lines, bypasses, drain and flare systems, control valves, and
instrumentation. The test set contains 23 scenarios.                    
                                                                      
The scenarios go beyond simple filtering of objects by type or          
properties. Some rules require only local conditions, such as the type  
of valve next to a given object, the first or last element near a       
diagram boundary, or the immediate neighbor of a piece of equipment.
Other rules take flow direction into account. For example, a flowmeter  
on a reflux line must be placed before the control valve and must be    
linked to the control lop of that valve. The remaining rules require    
path analysis: reachability between process systems, detection of       
bypass and recycle branches with a clear separation between the main    
flow and the bypass, and consistency of control lops and protective     
functions, such as the anti-surge loop of a compressor or temperature   
control downstream of a heat exchanger. Table~\ref{tab:rule-examples}   
lists several characteristic examples.                                  
                                                                      
\begin{table}[h]
\centering
\caption{Examples of rules from the test set.}
\label{tab:rule-examples}
\begin{tabular}{llp{8.5cm}}
\toprule
Rule & Check type & Content \\
\midrule
Rule 4   & Boundary, adjacency & The last element before the boundary is a shut-off or control valve \\
Rule 14  & Directed order      & On a reflux line, the flowmeter precedes the FCV and is linked to its control loop \\
Rule 136 & Path, control loop  & A centrifugal compressor has an anti-surge loop \\
Rule 665 & Forbidden bypass    & A fire-isolation ESV has no bypass path around it \\
\bottomrule
\end{tabular}
\end{table}

\subsubsection{Evaluation procedure}                                       

Each scenario is executed independently, and the state of the diagram   
is reset before every test. For each scenario we record three stages:
successful generation of the code, successful execution of the code,    
and a match between the obtained result and the reference answer.
Correctness is determined by the output of running the generated        
program on a fixed graph model, not by judging the wording of the
response.                                                               
                                                              
\subsubsection{Results}                                                    
                                                              
Table~\ref{tab:results} reports the evaluation results. All tested
models produced syntactically correct code for every scenario and
executed it successfully. The differences appear only at the level of   
semantic correctness. The best model solved all 23 scenarios, while
the two other models each failed on one scenario.                     
                                                                      
\begin{table}[h]
\centering
\caption{Accuracy on the test set of 23 scenarios.}
\label{tab:results}
\begin{tabular}{lrrrr}
\toprule
Model & Generated & Executed & Correct & Accuracy \\
\midrule
Qwen3.5-397B-A17B-FP8 & 23 & 23 & 23 & 100.00\,\% \\
Qwen3.6-35B-A3B       & 23 & 23 & 22 & 95.65\,\% \\
DeepSeek-V4-Pro       & 23 & 23 & 22 & 95.65\,\% \\
\bottomrule
\end{tabular}
\end{table}                                                   

\subsubsection{Oil-treatment PFD-to-P\&ID case study}
\label{subsec:pfd-to-pid-case-study}

The second experiment evaluates whether the proposed interaction layer can modify a valid PFD rather than only check an existing P\&ID. The input represents an oil-treatment installation and contains two centrifugal pumps, four heat exchangers, a gas separator, a water separator, a pipe tee, existing isolation valves, and off-page connectors. Its node-link representation contains 29 nodes and 28 directed process edges.

\paragraph{Rule set.}
The enrichment requirements were derived from a P\&ID project standard, Hydraulic Institute pump guidance, and API RP 12J separator practice \cite{KLM_PID_Standard,HydraulicInstitutePumpFAQ,API12J}. The following thirteen atomic rules were applied in the case study
(Table~\ref{tab:pfd-to-pid-rules}).

\begin{table*}[!t]
\centering
\caption{\textbf{PFD-to-P\&ID enrichment rules used in the oil-treatment case study.}}
\label{tab:pfd-to-pid-rules}
\renewcommand{\arraystretch}{1.15}
\setlength{\tabcolsep}{5pt}
\begin{tabular}{p{0.12\textwidth} p{0.22\textwidth} p{0.58\textwidth}}
\hline
\textbf{Rule} & \textbf{Equipment group} & \textbf{Requirement} \\
\hline
R1  & Centrifugal pump & Provide an isolation valve on the suction line. \\
R2  & Centrifugal pump & Provide a strainer upstream of the suction nozzle. \\
R3  & Centrifugal pump & Provide a pressure indicator on the discharge line between the pump nozzle and the check valve. \\
R4  & Centrifugal pump & Provide a check valve on the discharge line. \\
R5  & Centrifugal pump & Provide an isolation valve downstream of the check valve. \\
R6  & Heat exchanger & Provide isolation valves on the inlet and outlet process lines. \\
R7  & Heat exchanger & Provide temperature indicators at the inlet and outlet. \\
R8  & Heat exchanger & Provide valved vent and drain connections. \\
R9  & Separator & Provide level instrumentation indicating high, normal, and low liquid levels. \\
R10 & Separator & Provide pressure and temperature instrumentation. \\
R11 & Separator & Provide a valved vent and a valved drain. \\
R12 & Separator & Provide a pressure-relief connection. \\
R13 & Outlet boundary & Provide an isolation valve before the off-page connector. \\
\hline
\end{tabular}
\end{table*}

The rules were selected from sources that represent complementary levels of engineering guidance. The KLM document was used as the primary source because it is a publicly available project engineering standard that provides explicit P\&ID requirements for equipment isolation, instrumentation, vents, drains, and pressure-relief connections. The Hydraulic Institute guidance was used to support pump-specific arrangements, particularly the placement of suction and discharge isolation valves and check valves. API Specification 12J was included as an industry-specific reference for oil and gas separators, making it directly relevant to the considered oil-treatment process. Together, these sources provide general P\&ID design guidance and equipment-specific requirements applicable to the evaluated diagram.

\paragraph{Rule application and verification.}
The rules were applied in four equipment-specific stages: pumps, heat
exchangers, separators, and outlet boundaries. At each stage, the agent
interpreted the applicable requirements and generated the corresponding
graph modifications. Existing objects were retained whenever they already
satisfied a requirement. The modified graph was then checked for the
presence, type, and placement of the required P\&ID elements.

After all four stages, the complete enriched graph was compared with an
independently implemented deterministic reference in terms of object
composition, directed process topology, and attachment relations. Inserted
objects received deterministic identifiers. When an existing process edge
was split, its stream type, fluid type, nominal diameter, and flow attributes
were preserved in the resulting edges.

\begin{figure*}[!t]
\centering
\small
\setlength{\tabcolsep}{3pt}
\renewcommand{\arraystretch}{1.05}

\begin{tabular}{
p{0.22\textwidth}
>{\centering\arraybackslash}p{0.37\textwidth}
>{\centering\arraybackslash}p{0.37\textwidth}
}
\textbf{Rule} & \textbf{Input fragment} & \textbf{Enriched fragment} \\

Rule~4: pump discharge check valve
&
\includegraphics[width=\linewidth]{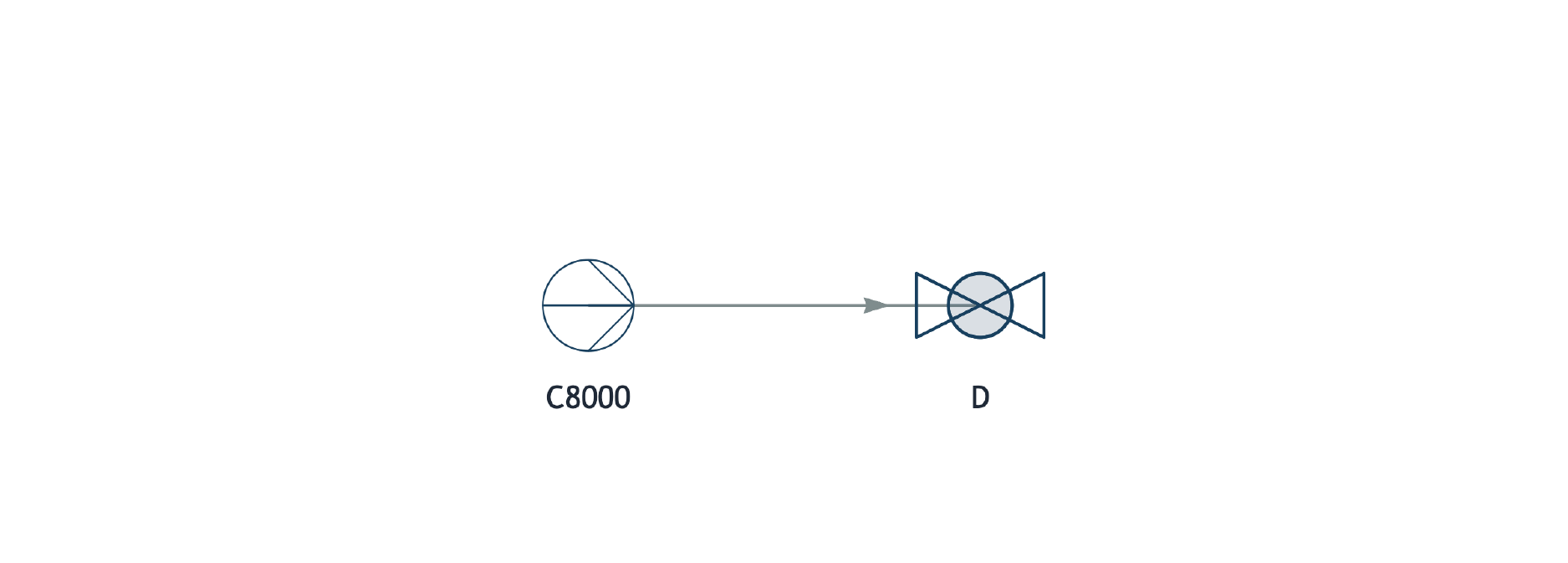}
&
\includegraphics[width=\linewidth]{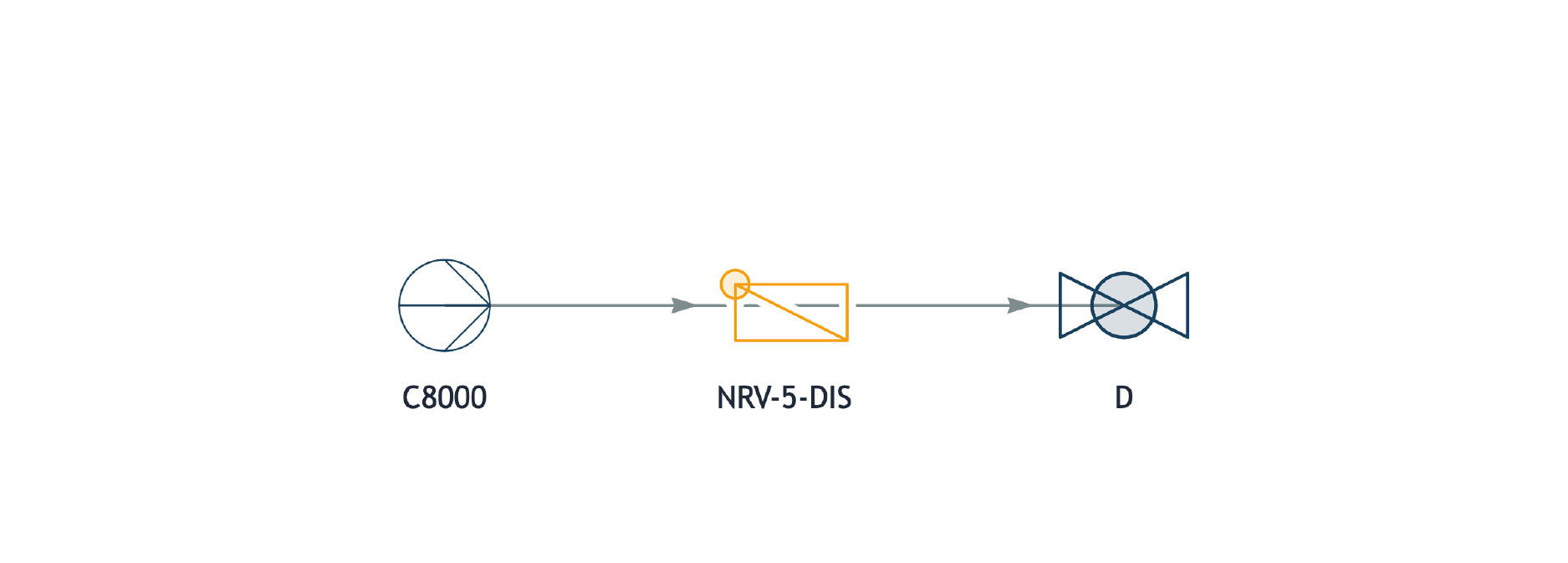}
\\[-0.2em]

Rule~2: pump suction strainer
&
\includegraphics[width=\linewidth]{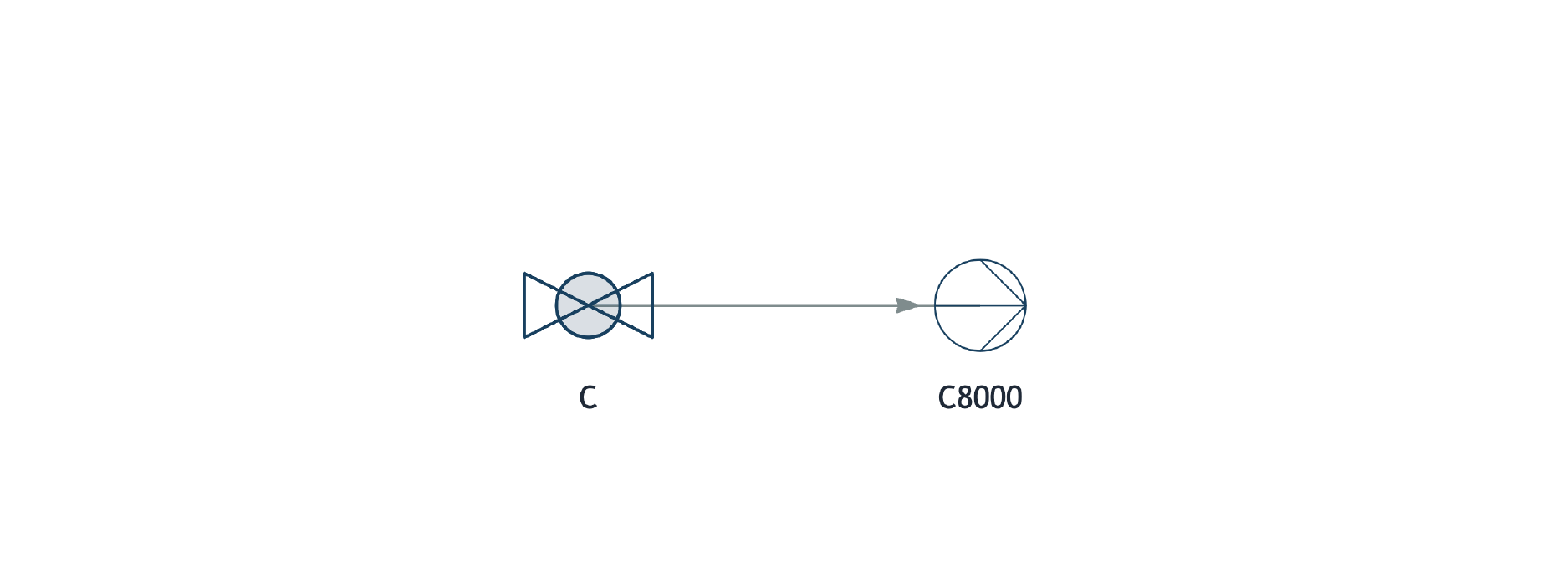}
&
\includegraphics[width=\linewidth]{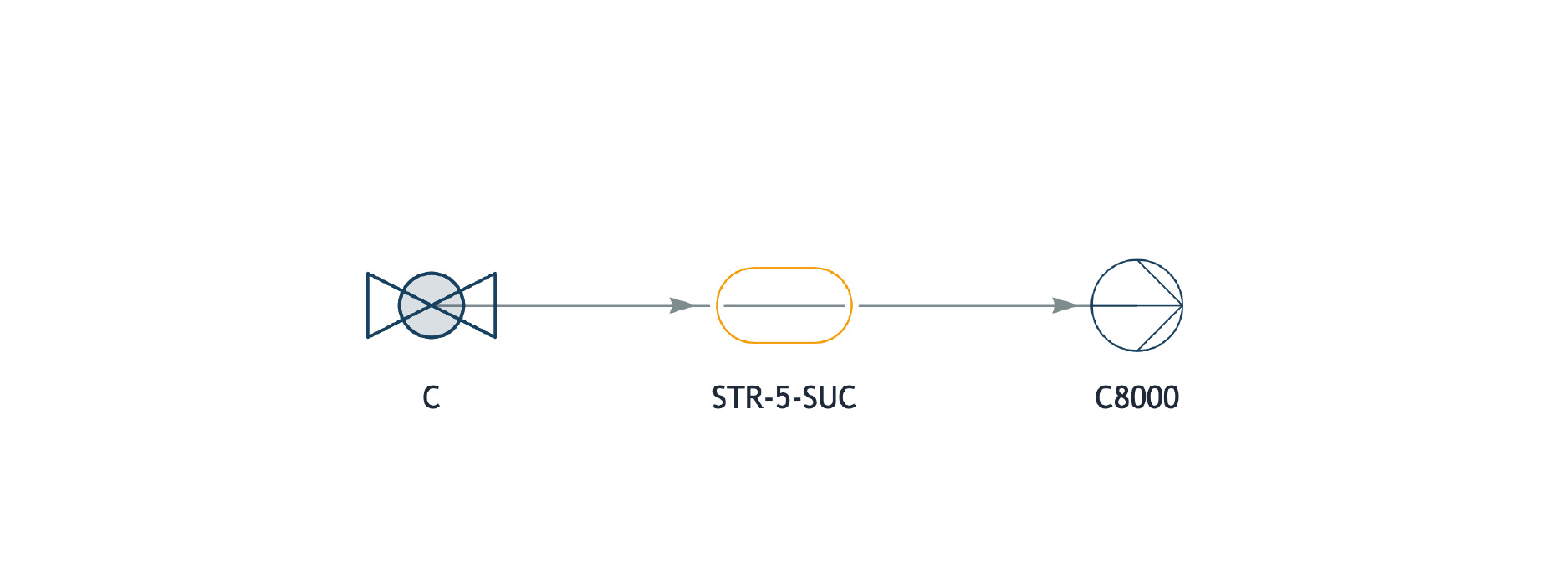}
\\[-0.2em]

Rule~3: pump discharge pressure indication
&
\includegraphics[width=\linewidth]{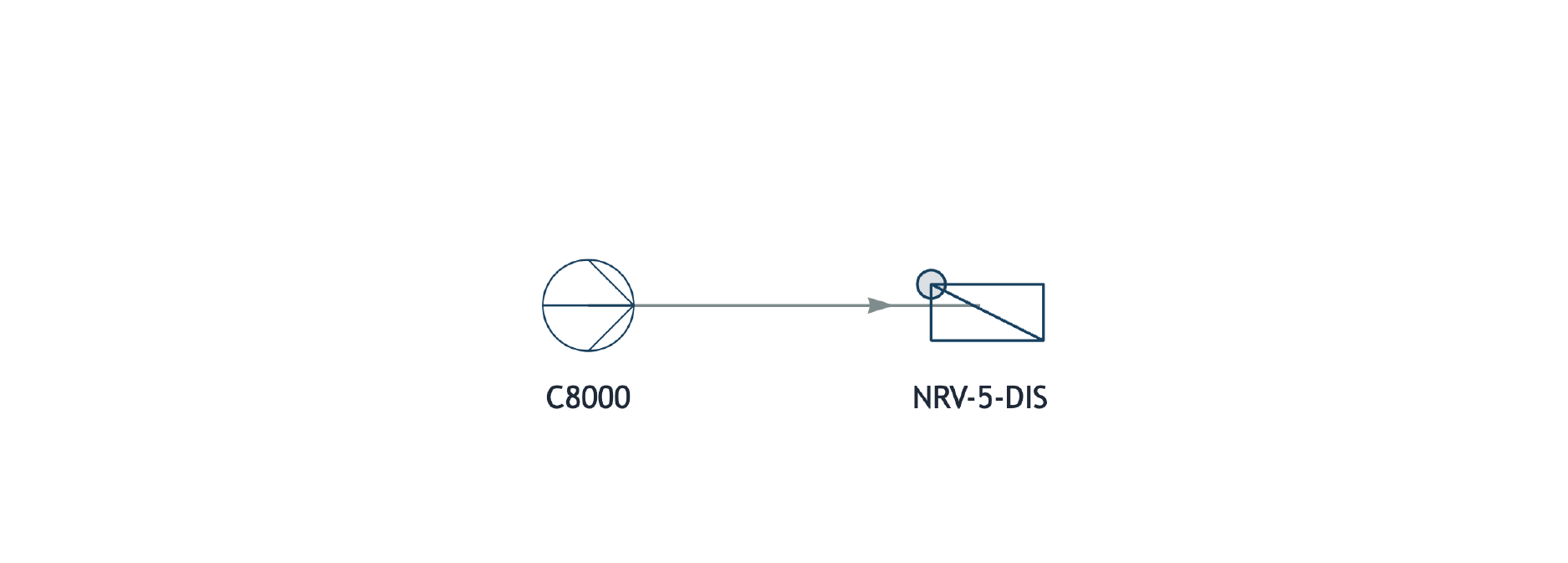}
&
\includegraphics[width=\linewidth]{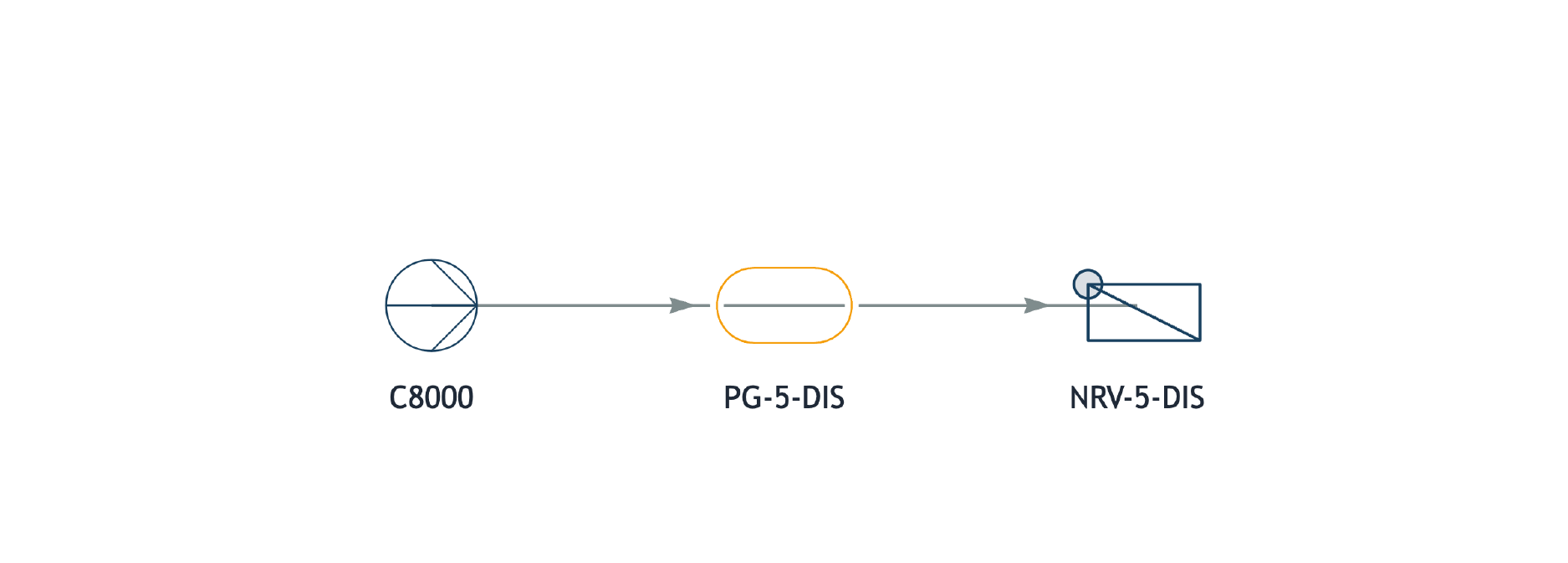}
\\[-0.2em]

Rule~7: heat-exchanger temperature indication
&
\includegraphics[width=\linewidth]{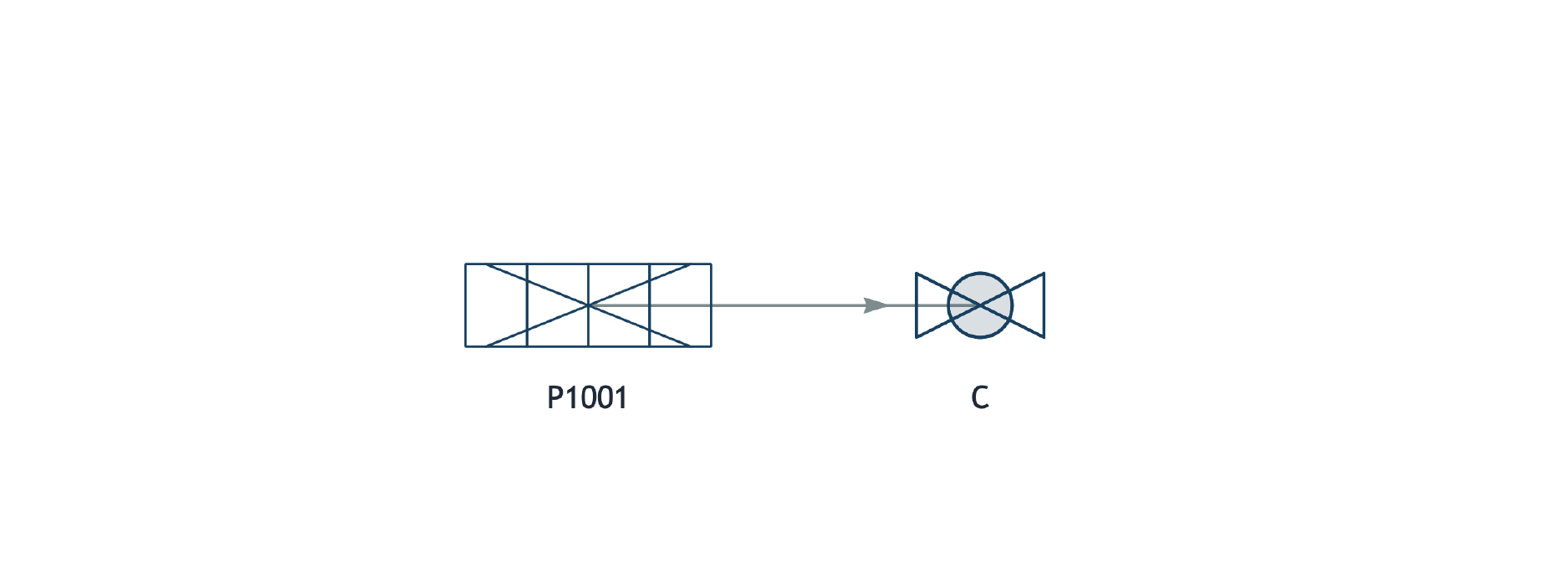}
&
\includegraphics[width=\linewidth]{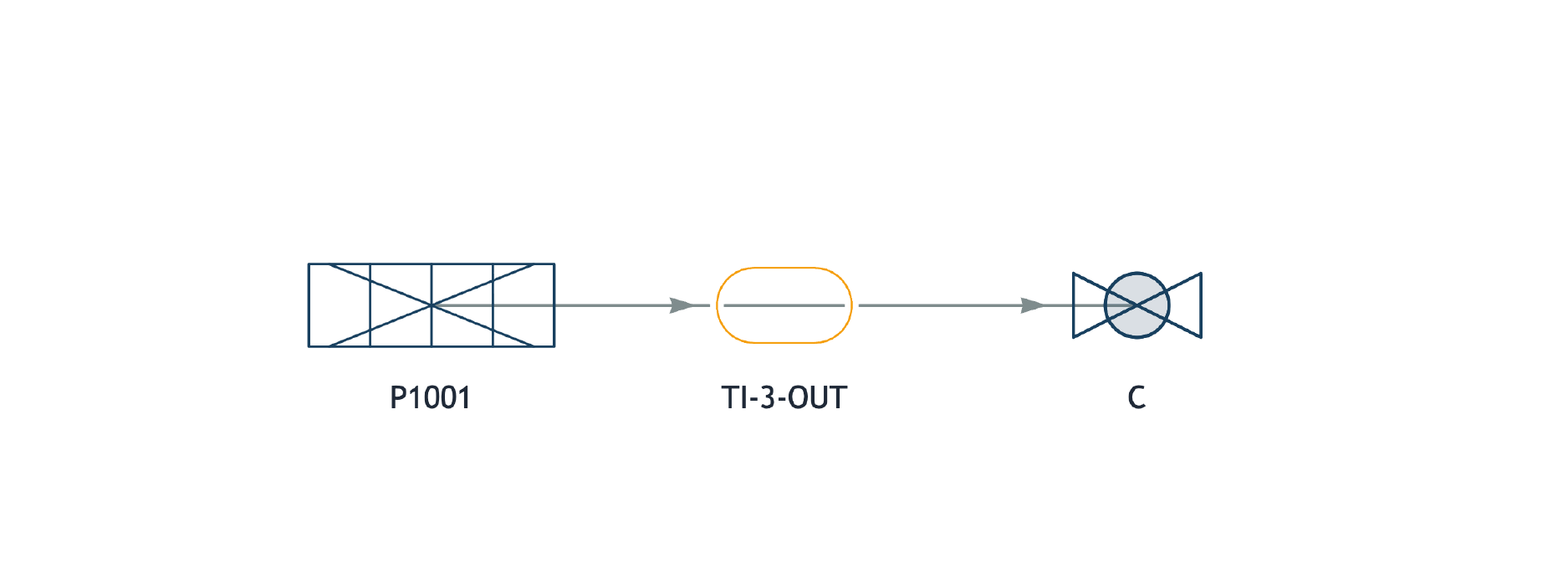}
\\[-0.2em]

Rule~12: separator pressure-relief connection
&
\includegraphics[width=\linewidth]{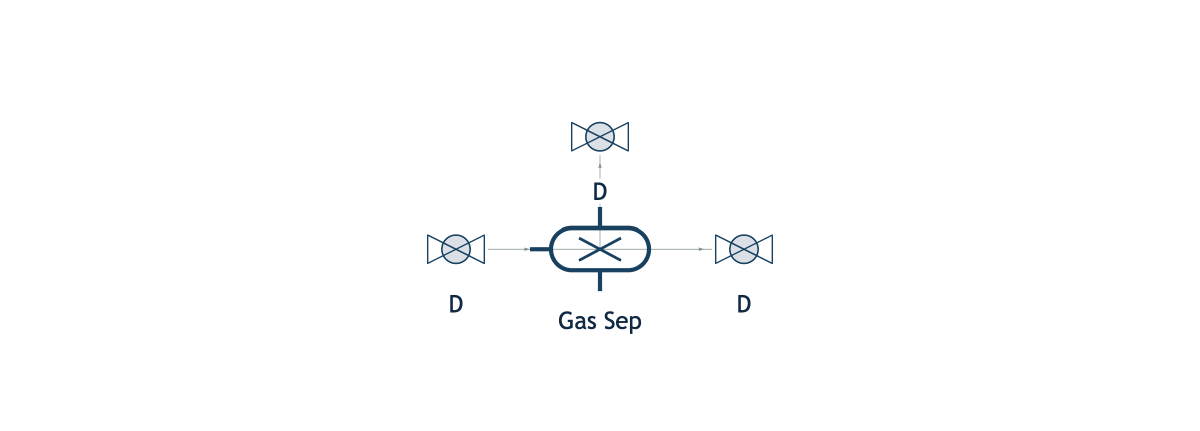}
&
\includegraphics[width=\linewidth]{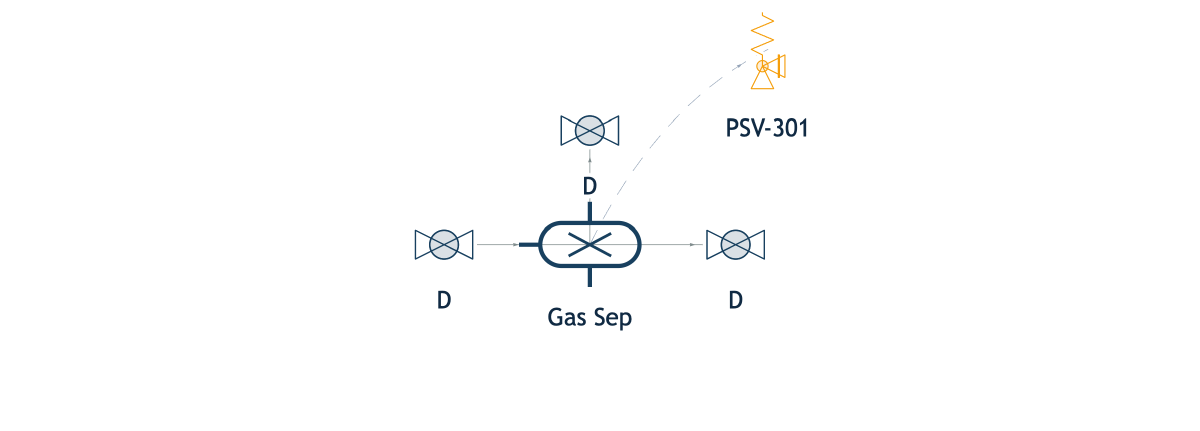}
\\[-0.2em]

Rule~11: separator valved drain connection
&
\includegraphics[width=\linewidth]{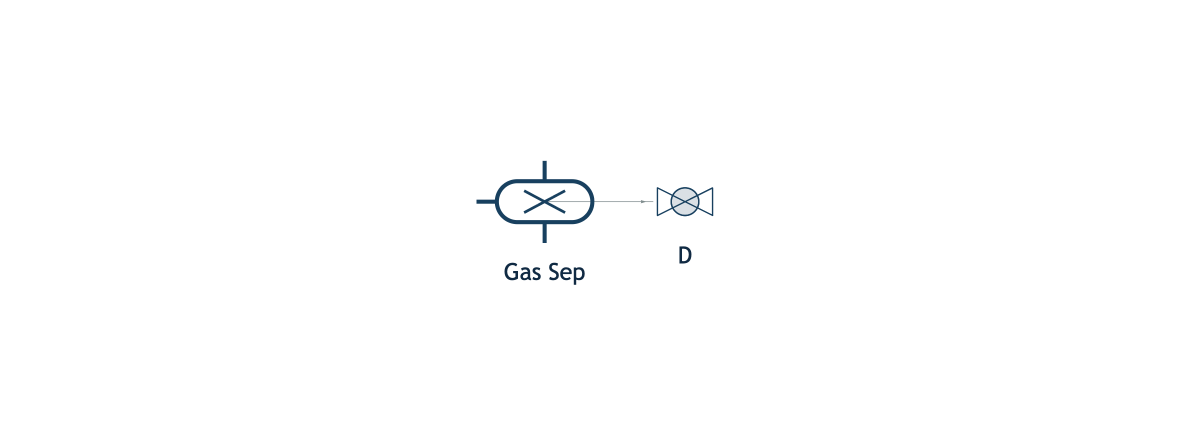}
&
\includegraphics[width=\linewidth]{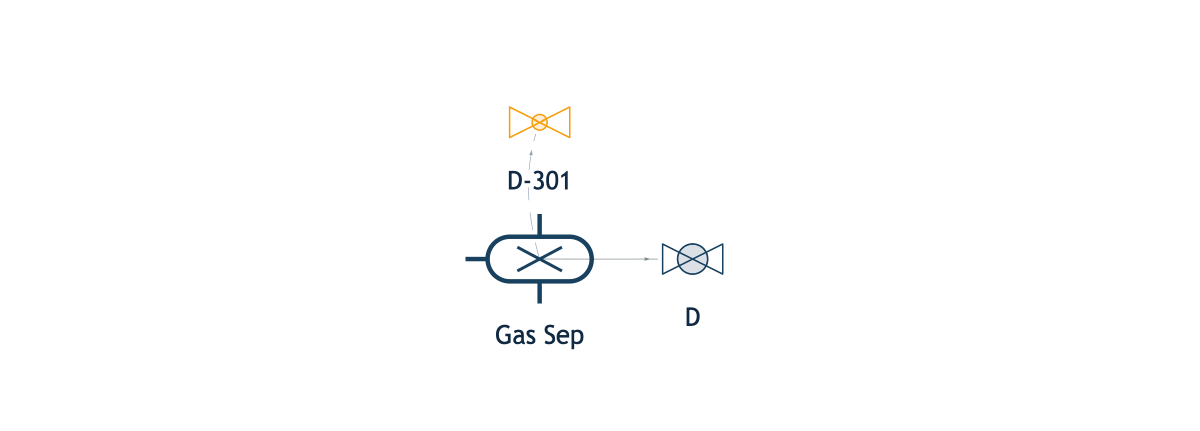}
\end{tabular}

\caption{\textbf{Representative rule-level transformations from PFD to
P\&ID.} Each row shows one atomic enrichment rule, with the input graph
fragment in the middle column and the corresponding enriched fragment in
the right column. Inserted objects are highlighted in yellow. Only the local
graph fragment relevant to each rule is shown. Node positions and graphical
line routes are generated automatically for readability and do not encode
additional engineering information.}
\label{fig:atomic-enrichment-examples}
\end{figure*}

Representative local effects of six enrichment rules are shown in
Figure ~\ref{fig:atomic-enrichment-examples}. Each example shows only the
local graph fragment required to inspect the corresponding rule.

\paragraph{Results.}
The LLM-generated programs introduced 38 P\&ID objects and reused 18 objects already present in the PFD. The resulting graph contained 67 nodes and 66 edges. Comparison with the deterministic reference found no missing or additional nodes, no semantic object-type mismatches, and no differences in either process topology or equipment-attachment relations.

\begin{table}[H]
\centering
\caption{End-to-end PFD-to-P\&ID enrichment results.}
\label{tab:pfd-pid-e2e-results}
\begin{tabular}{lr}
\toprule
Metric & Value \\
\midrule
Input graph & 29 nodes, 28 edges \\
New P\&ID objects & 38 \\
Existing objects reused & 18 \\
Output graph & 67 nodes, 66 edges \\
Missing / extra nodes & 0 / 0 \\
Semantic object-type mismatches & 0 \\
Missing / extra process edges & 0 / 0 \\
Missing / extra attachment relations & 0 / 0 \\
\bottomrule
\end{tabular}
\end{table}

\FloatBarrier

\subsubsection{Limitations}

The evaluation was conducted in two controlled settings: a benchmark of 23 rule-checking scenarios and a PFD-to-P\&ID transformation for an oil-treatment process. Although the generated graph matched the reference structure, this result is limited to the selected engineering rules, equipment types, and process topology. Additional experiments are required to assess the generalizability of the approach to other processes and project-specific requirements. Moreover, the experiment evaluates graph-level consistency rather than the completeness of a construction-ready P\&ID. Detailed safety studies, equipment sizing, and final engineering validation remain outside the scope of this work.
The evaluation focuses on semantic and topological correctness; graphical
sheet layout is outside its scope and is not included in the reported
comparisons.

\section{Conclusion}

This work presented a Full Cycle AI pipeline that automates both optimal PFD synthesis and its subsequent transformation into a validated P\&ID. The results demonstrate that a hybrid GA/LLM approach generates optimal, rule-compliant PFD topologies, while the LLM-based transformation agent, constrained by the SDK, reliably produces source-grounded P\&ID modifications. By integrating these two stages within a single workflow, the proposed framework indeed may reduce manual engineering effort and accelerate the exploration of alternative design configurations. Crucially, the built-in validation mechanisms against domain-specific rules and graph structures ensure that the generated outputs are not only novel but also practically deployable in real-world engineering settings. 

\printbibliography

\clearpage
\appendix
\section{Equations behind flow calculations}
\label{app:equations}

The following equations describe the calculations performed by the \texttt{NodePhysics} dispatcher.  
All flows are assumed at steady state.  
Pressures in Pa, temperatures in °C or K as indicated.

\subsection{Gas separator}
A gas separator removes a fraction $\varepsilon$ of the gas phase from the incoming oil–water–gas mixture. The removal efficiency depends on pressure and temperature:
\begin{equation}
    \varepsilon(P,T) = \varepsilon_{\text{base}} \left[1 + \alpha_p \frac{P-P_{\text{ref}}}{P_{\text{ref}}}\right] 
    \left[1 - \alpha_t (T-T_{\text{ref}})\right],
\end{equation}
with $0 \le \varepsilon \le 0.99$. The outlet liquid stream has updated phase fractions:
\begin{align}
    y_{\text{gas,out}} &= \frac{y_{\text{gas,in}}(1-\varepsilon)}{1 - \varepsilon\, y_{\text{gas,in}}}, \\
    y_{\text{oil,out}} &= \frac{y_{\text{oil,in}}}{1 - \varepsilon\, y_{\text{gas,in}}}, \\
    y_{\text{water,out}} &= \frac{y_{\text{water,in}}}{1 - \varepsilon\, y_{\text{gas,in}}}.
\end{align}
The removed gas forms a separate stream (pure gas). A pressure drop $\Delta P_{\text{sep}}$ (e.g., 50 kPa) is applied to both outlets. After updating the phase fractions, the mixture thermophysical properties (density and heat capacity) are recalculated from the new composition.

\subsection{Water separator}
A water separator splits the incoming stream into an oil‑rich stream and a water‑rich stream using fixed separation efficiencies. Let:
\begin{itemize}
    \item $X_{w|o}$ – mass fraction of water remaining in the oil outlet,
    \item $X_{o|w}$ – mass fraction of oil remaining in the water outlet.
\end{itemize}
The total mass flow $\dot{m}_{\text{in}}$ and inlet oil/water mass fractions $X_o$, $X_w$ are known (gas is assumed to be negligible, although any residual gas fraction is propagated with the oil outlet.). Solving the material balance:
\begin{align}
    \dot{m}_o &= \dot{m}_{\text{in}}\, \frac{X_o (1-X_{o|w}) - X_w X_{o|w}}
    {(1-X_{w|o})(1-X_{o|w}) - X_{w|o} X_{o|w}}, \\
    \dot{m}_w &= \dot{m}_{\text{in}} - \dot{m}_o,
\end{align}
where $\dot{m}_o$ and $\dot{m}_w$ are the mass flow rates of the oil and water outlets. Their phase fractions are:
\begin{align}
    \text{Oil outlet:}&\quad y_{\text{oil,o}} = 1 - X_{w|o},\quad y_{\text{water,o}} = X_{w|o},\quad y_{\text{gas,o}} = 0,\\
    \text{Water outlet:}&\quad y_{\text{oil,w}} = X_{o|w},\quad y_{\text{water,w}} = 1 - X_{o|w},\quad y_{\text{gas,w}} = 0.
\end{align}
Both outlets experience a pressure drop $\Delta P_{\text{wsep}}$ (typically 20 kPa).

\subsection{Pump}
The pump adds mechanical energy to the fluid. The head $H$ (in meters of fluid) and efficiency $\eta$ are quadratic functions of the volumetric flow rate $Q$ (\si{m^3/h}):
\begin{align}
    H(Q) &= a_2 Q^2 + a_1 Q + a_0, \\
    \eta(Q) &= b_2 Q^2 + b_1 Q + b_0,
\end{align}
with coefficients supplied from a pump catalogue. The pressure rise is:
\begin{equation}
    \Delta P = \rho \, g \, H(Q),
\end{equation}
where $\rho$ is the fluid density (\si{kg/m^3}) and $g = 9.81\ \si{m/s^2}$. The outlet pressure becomes $P_{\text{out}} = P_{\text{in}} + \Delta P$.  
A violation is raised if:
\begin{itemize}
    \item $P_{\text{in}} < P_{\text{min,inlet}}$ (cavitation risk),
    \item $\eta(Q) < 0.7\,\eta_{\max}$ (operating far from best efficiency point, where $\eta_{\max}$denotes the maximum value of the efficiency curve over the operating flow range).
\end{itemize}

\subsection{Heat exchanger}
The heat exchanger heats (or cools) the fluid using a hot utility at constant temperature $T_h$. The transferred heat is controlled by a factor $F_k$ (\si{W/K}) and the actual heat capacity rate $\dot{m} c_p$:
\begin{equation}
    T_{\text{out}} = T_{\text{in}} + \frac{F_k (T_h - T_{\text{in}})}{\dot{m} c_p},
\end{equation}
provided the inlet pressure does not exceed $P_{\text{max,inlet}}$. A small pressure drop $\Delta P_{\text{hx}}$ (e.g., 10 kPa) is subtracted from the outlet pressure.

\subsection{Pipe tee (splitter)}
A pipe tee splits the flow equally among all outgoing branches. For $N$ outlets:
\begin{equation}
    \dot{m}_{\text{branch}} = \frac{\dot{m}_{\text{in}}}{N}, \quad 
    Q_{\text{branch}} = \frac{Q_{\text{in}}}{N}, \quad
    P_{\text{branch}} = P_{\text{in}} \cdot (1 - \delta_{\text{tee}}),
\end{equation}
where $\delta_{\text{tee}} = 0.02$ (2\% pressure loss). For simplicity each outgoing branch inherits the inlet thermodynamic state (temperature, composition, density, and heat capacity), while only the flow rate and pressure are modified.

\subsection{Ball valve}

The ball valve is modeled as an ideal isolation element that does not modify the flow properties. The outlet stream therefore preserves the inlet pressure, temperature, flow rate, and phase composition. Its only function is to verify that the operating pressure does not exceed the valve's maximum allowable working pressure:
\begin{equation}
    P_{\mathrm{in}} \le P_{\mathrm{max,working}}.
\end{equation}

If this condition is violated, the corresponding node is marked with a pressure violation. No pressure drop across the valve is assumed.

\clearpage
\section{Algorithms}
\label{app:algorithms}
This appendix contains algorithmic descriptions of the Python solver and the loss calculation functions.

\vspace{0.5cm}

\noindent\textbf{Algorithm B.1: Graph-Based Process Flow Solver}

\begin{algorithmic}
\Require Graph $G$, initial flow $F_{\text{in}}$
\Ensure Enriched graph $G$

\State Build edge maps $E_{\text{out}}$, $E_{\text{in}}$, find inlet node $n_{\text{in}}$
\State Assign initial flow properties to an inlet node: $F[n_{\text{in}}] \gets F_{\text{in}}$
\State Initialize propagation queue with the inlet node and create empty list of processed nodes: $Q \gets [n_{\text{in}}]$, $P \gets \emptyset$

\While{$Q \neq \emptyset$}
    \State Pop $n$ from $Q$
    \If{$n \in P$} \State \textbf{continue} \EndIf

    \State Collect incoming streams of node $n$: $E_{\text{in}}(n) = \{e_1,\dots,e_k\}$
    \If{$k > 1$}
        \State Dealing with several incoming edges 
        \State Merge flows: $\{F_i\} \gets \text{flows from } E_{\text{in}}(n)$
        \State Recalculate merged flow properties: $F \gets \text{Mix}(F_1,\dots,F_k)$
    \Else
        \State Dealing with one incoming edge
        \State $F \gets F[n]$
    \EndIf

    \State Get node $n$ equipment type: $\text{type} \gets \text{label}(n)$
    \State Perform equipment specific calculation - get updated flow properties $F_{\text{out}}$ and check for violations $V$: 
    \State $(F_{\text{out}}, V) \gets \text{NodePhysics}(F, \text{type})$
    \State Store violations: $V_{\text{store}}[n] \gets V$

    \For{each $e \in E_{\text{out}}(n)$}
        \State For each outgoing edge, propagate flow to downstream node
        \State $F_e \gets F_{\text{out}}[\text{stream\_type}(e)]$
        \State $e.\text{flow} \gets F_e$
        \State $m \gets e.\text{to}$
        \State $F[m] \gets F_e$
        \State $Q.\text{append}(m)$
    \EndFor
    
    \State Add node $n$ to list of processed nodes: $P \gets P \cup \{n\}$
\EndWhile

\State \Return $G$

\end{algorithmic}

\vspace{0.5cm}


\noindent\textbf{Algorithm B.2: Loss Evaluation from generated chromosome}

\begin{algorithmic}[1]
\Require
Chromosome $\mathbf{s}$,
equipment database $D_{eq}$,
inlet flow properties $F_{in}$,
target outlet properties $F_{out}$,
solver function $\mathcal{S}$,
mapping dictionary $M$

\Ensure
Fitness value $f$

\State Construct graph:
\[
G \gets \textsc{BuildGraphFromChromosome}
(\mathbf{s}, M, D_{eq})
\]

\State Pass received graph to the solver:
\[
G \gets \mathcal{S}(G, F_{in})
\]

\State Compute total equipment cost:
\[
C_{\text{tot}} \gets
\textsc{ComputeCostFromGraph}(G)
\]

\State Compute number of solver violations:
\[
N_{\text{solver}}
\gets
\sum_{n \in G_{\text{nodes}}}
|\text{violations}(n)|
\]

\State Compute topology violations for ball valve rule:
\[
N_{\text{topo}}
\gets
\textsc{CheckBallValveRule}(G)
\]

\State Compute outlet deviation penalty:
\[
P_{\text{dev}}
\gets
\textsc{OutletDeviationPenalty}
(G, F_{out})
\]

\State Compute the penalty for ball valves following each other in a row:
\[
P_{\text{bv}}
\gets
\textsc{BallValveInaRowPenalty}
(G)
\]

\State Compute core equipment presence penalty:
\[
P_{\text{core}}
\gets
\textsc{CoreEquipment PresencePenalty}
(G)
\]

\State Compute total objective:
\[
J =
C_{\text{tot}}
+
10^5 N_{\text{solver}}
+
10^6 N_{\text{topo}}
+
P_{\text{dev}}
+
P_{\text{bv}}
+
P_{\text{core}}
\]

\State Compute fitness:
\[
f \gets -J
\]

\Return $f$

\end{algorithmic}








\clearpage
\section{Agents interaction in MAS}
\label{app:agents interaction}

This appendix contains the MAS agents' interaction process for the last two iterations (for model Qwen3.6-35B-A3B).

\subsection{Iteration 4}

\begin{optimizerbox}

\begin{lstlisting}[language=json]
{
"reasoning": "Fixed the topology connection errors by combining node creation and relationship creation into single sequential queries. This ensures that each newly created node is immediately connected to the previous one within the same query execution context, eliminating the UUID mismatch issue caused by separate CREATE and MATCH statements. The topology follows the optimal path: Inlet -> BV -> PipeTee (splits flow) -> BV -> MP Separator (gas removal) -> BV -> SETTLER_3 (water removal) -> BV -> R7000 Pump (pressure boost) -> BV -> P1002 Heat Exchanger (temperature boost) -> BV -> Outlet. Branches correctly route waste, gas, and water streams. All equipment properties match tool outputs, and inlet/outlet handling follows constraints.",
"cypher_queries": ["MATCH (inlet {_id: '5161c1ac-e287-49f7-b658-4128d42db9e1'}) CREATE (inlet)-[:CONNECTED {_type:'Pipe', stream_type:'main'}]->(bv1:BallValve {id: randomUUID(), _id: randomUUID(), name: 'D', _label: 'BallValve', maxWorkingPressure: 600000, cost: 400})", "MATCH (bv1) CREATE (bv1)-[:CONNECTED {_type:'Pipe', stream_type:'main'}]->(pt:PipeTee {id: randomUUID(), _id: randomUUID(), name: 'PT', _label: 'PipeTee', cost: 2000})", "MATCH (pt) CREATE (pt)-[:CONNECTED {_type:'Pipe', stream_type:'main'}]->(bv2:BallValve {id: randomUUID(), _id: randomUUID(), name: 'D', _label: 'BallValve', maxWorkingPressure: 600000, cost: 400})", "MATCH (pt) CREATE (pt)-[:CONNECTED {_type:'Pipe', stream_type:'main'}]->(bvX:BallValve {id: randomUUID(), _id: randomUUID(), name: 'D', _label: 'BallValve', maxWorkingPressure: 600000, cost: 400})", "MATCH (bv2) CREATE (bv2)-[:CONNECTED {_type:'Pipe', stream_type:'main'}]->(mp:GasSeparator {id: randomUUID(), _id: randomUUID(), name: 'MP Separator', _label: 'GasSeparator', gas_removal_eff: 0.8, pressure_drop: 15000, cost: 5000})", "MATCH (mp) CREATE (mp)-[:CONNECTED {_type:'Pipe', stream_type:'main'}]->(bv3:BallValve {id: randomUUID(), _id: randomUUID(), name: 'D', _label: 'BallValve', maxWorkingPressure: 600000, cost: 400})", "MATCH (mp) CREATE (mp)-[:CONNECTED {_type:'Pipe', stream_type:'gas'}]->(bv4:BallValve {id: randomUUID(), _id: randomUUID(), name: 'D', _label: 'BallValve', maxWorkingPressure: 600000, cost: 400})", "MATCH (bv3) CREATE (bv3)-[:CONNECTED {_type:'Pipe', stream_type:'main'}]->(ws:WaterSeparator {id: randomUUID(), _id: randomUUID(), name: 'SETTLER_3', _label: 'WaterSeparator', X_w_o: 0.01, X_o_w: 0.0001, pressure_drop: 5000, cost: 4300})", "MATCH (ws) CREATE (ws)-[:CONNECTED {_type:'Pipe', stream_type:'main'}]->(bv5:BallValve {id: randomUUID(), _id: randomUUID(), name: 'D', _label: 'BallValve', maxWorkingPressure: 600000, cost: 400})", "MATCH (ws) CREATE (ws)-[:CONNECTED {_type:'Pipe', stream_type:'water'}]->(bv6:BallValve {id: randomUUID(), _id: randomUUID(), name: 'D', _label: 'BallValve', maxWorkingPressure: 600000, cost: 400})", "MATCH (bv5) CREATE (bv5)-[:CONNECTED {_type:'Pipe', stream_type:'main'}]->(pump:ReciprocatingPump {id: randomUUID(), _id: randomUUID(), name: 'R7000', _label: 'ReciprocatingPump', designVolumeFlowRate: 500, cost: 4800, min_inlet_pressure: 70000, head_coeffs: [-5.6e-05, 0.0, 24.0], eff_coeffs: [-4.8e-06, 0.0027, 0.0]})", "MATCH (pump) CREATE (pump)-[:CONNECTED {_type:'Pipe', stream_type:'main'}]->(bv7:BallValve {id: randomUUID(), _id: randomUUID(), name: 'D', _label: 'BallValve', maxWorkingPressure: 600000, cost: 400})", "MATCH (bv7) CREATE (bv7)-[:CONNECTED {_type:'Pipe', stream_type:'main'}]->(hxe:PlateHeatExchanger {id: randomUUID(), _id: randomUUID(), name: 'P1002', _label: 'PlateHeatExchanger', designHeatTransferArea_m2: 46.8, Fk: 23280, cost: 6100, max_inlet_pressure: 495000, T_h: 80})", "MATCH (hxe) CREATE (hxe)-[:CONNECTED {_type:'Pipe', stream_type:'main'}]->(bv8:BallValve {id: randomUUID(), _id: randomUUID(), name: 'D', _label: 'BallValve', maxWorkingPressure: 600000, cost: 400})", "MATCH (bv8) CREATE (bv8)-[:CONNECTED {_type:'Pipe', stream_type:'main'}]->(outlet:FlowOutPipeOffPageConnector {id: randomUUID(), _id: randomUUID(), name: 'Outlet', _label: 'FlowOutPipeOffPageConnector', cost: 0})"], 
"summary":{
    "estimated_equipment_count": 15, 
    "pump_count": 1,
    "heater_count": 1,
    "separator_count": 2,
    "branch_count": 3
    }
}
\end{lstlisting}
\end{optimizerbox}

\begin{cypherbox}

\begin{lstlisting}[language=json]
{
"validation_status": false,
"fixed_queries": [],
"validation_details":{
    "syntax_errors_found": 0,
    "queries_fixed": 0,
    "reasoning": "Queries 2 through 15 violate the variable scoping rule. Each query references a node variable (e.g., bv1, pt, bv2, mp, ws, pump, bv7, bv8) that was created in a previous query. The validation rules explicitly state: 'Every Cypher statement must be self-contained. Variables may not be reused across separate statements unless they are MATCHed in the same statement.' Since these are provided as separate statements, each must independently match its target nodes (preferably by _id) or be combined into a single self-contained statement. Additionally, because _id values are generated dynamically via randomUUID() in Query 1, subsequent queries cannot know the exact UUIDs to match by _id without external state management, making the current sequential structure invalid for independent execution."
    }
}
\end{lstlisting}

\end{cypherbox}

\subsection{Iteration 5}

\begin{optimizerbox}

\begin{lstlisting}[language=json]
{
"reasoning": "Combined all node creation and relationship establishment into a single self-contained Cypher query to strictly satisfy the variable scoping rule. By executing all CREATE clauses and relationship mappings in one statement, we eliminate cross-query variable dependencies and ensure independent executability. The query matches the existing inlet node, generates all equipment nodes with dynamic UUIDs via randomUUID(), and connects them according to the optimal topology: Inlet -> BV -> PipeTee (splits flow) -> BV -> MP Separator (gas removal) -> BV -> SETTLER_3 (water removal) -> BV -> R7000 Pump (pressure boost) -> BV -> P1002 Heat Exchanger (temperature boost) -> BV -> Outlet. Branches correctly route waste, gas, and water streams.",
"cypher_queries": ["MATCH (inlet {_id: '5161c1ac-e287-49f7-b658-4128d42db9e1'}) CREATE (bv1:BallValve {id: randomUUID(), _id: randomUUID(), name: 'D', _label: 'BallValve', maxWorkingPressure: 600000, cost: 400}), (pt:PipeTee {id: randomUUID(), _id: randomUUID(), name: 'PT', _label: 'PipeTee', cost: 2000}), (bv2:BallValve {id: randomUUID(), _id: randomUUID(), name: 'D', _label: 'BallValve', maxWorkingPressure: 600000, cost: 400}), (bvX:BallValve {id: randomUUID(), _id: randomUUID(), name: 'D', _label: 'BallValve', maxWorkingPressure: 600000, cost: 400}), (mp:GasSeparator {id: randomUUID(), _id: randomUUID(), name: 'MP Separator', _label: 'GasSeparator', gas_removal_eff: 0.8, pressure_drop: 15000, cost: 5000}), (bv3:BallValve {id: randomUUID(), _id: randomUUID(), name: 'D', _label: 'BallValve', maxWorkingPressure: 600000, cost: 400}), (bv4:BallValve {id: randomUUID(), _id: randomUUID(), name: 'D', _label: 'BallValve', maxWorkingPressure: 600000, cost: 400}), (ws:WaterSeparator {id: randomUUID(), _id: randomUUID(), name: 'SETTLER_3', _label: 'WaterSeparator', X_w_o: 0.01, X_o_w: 0.0001, pressure_drop: 5000, cost: 4300}), (bv5:BallValve {id: randomUUID(), _id: randomUUID(), name: 'D', _label: 'BallValve', maxWorkingPressure: 600000, cost: 400}), (bv6:BallValve {id: randomUUID(), _id: randomUUID(), name: 'D', _label: 'BallValve', maxWorkingPressure: 600000, cost: 400}), (pump:ReciprocatingPump {id: randomUUID(), _id: randomUUID(), name: 'R7000', _label: 'ReciprocatingPump', designVolumeFlowRate: 500, cost: 4800, min_inlet_pressure: 70000, head_coeffs: [-5.6e-05, 0.0, 24.0], eff_coeffs: [-4.8e-06, 0.0027, 0.0]}), (bv7:BallValve {id: randomUUID(), _id: randomUUID(), name: 'D', _label: 'BallValve', maxWorkingPressure: 600000, cost: 400}), (hxe:PlateHeatExchanger {id: randomUUID(), _id: randomUUID(), name: 'P1002', _label: 'PlateHeatExchanger', designHeatTransferArea_m2: 46.8, Fk: 23280, cost: 6100, max_inlet_pressure: 495000, T_h: 80}), (bv8:BallValve {id: randomUUID(), _id: randomUUID(), name: 'D', _label: 'BallValve', maxWorkingPressure: 600000, cost: 400}), (outlet:FlowOutPipeOffPageConnector {id: randomUUID(), _id: randomUUID(), name: 'Outlet', _label: 'FlowOutPipeOffPageConnector', cost: 0}), (inlet)-[:CONNECTED {_type:'Pipe', stream_type:'main'}]->(bv1), (bv1)-[:CONNECTED {_type:'Pipe', stream_type:'main'}]->(pt), (pt)-[:CONNECTED {_type:'Pipe', stream_type:'main'}]->(bv2), (pt)-[:CONNECTED {_type:'Pipe', stream_type:'main'}]->(bvX), (bv2)-[:CONNECTED {_type:'Pipe', stream_type:'main'}]->(mp), (mp)-[:CONNECTED {_type:'Pipe', stream_type:'main'}]->(bv3), (mp)-[:CONNECTED {_type:'Pipe', stream_type:'gas'}]->(bv4), (bv3)-[:CONNECTED {_type:'Pipe', stream_type:'main'}]->(ws), (ws)-[:CONNECTED {_type:'Pipe', stream_type:'main'}]->(bv5), (ws)-[:CONNECTED {_type:'Pipe', stream_type:'water'}]->(bv6), (bv5)-[:CONNECTED {_type:'Pipe', stream_type:'main'}]->(pump), (pump)-[:CONNECTED {_type:'Pipe', stream_type:'main'}]->(bv7), (bv7)-[:CONNECTED {_type:'Pipe', stream_type:'main'}]->(hxe), (hxe)-[:CONNECTED {_type:'Pipe', stream_type:'main'}]->(bv8), (bv8)-[:CONNECTED {_type:'Pipe', stream_type:'main'}]->(outlet)"],
"summary":{
    "estimated_equipment_count": 15, 
    "pump_count": 1,
    "heater_count": 1,
    "separator_count": 2,
    "branch_count": 3
    }
}
\end{lstlisting}
\end{optimizerbox}

\begin{cypherbox}

\begin{lstlisting}[language=json]
{
"validation_status": true,
"fixed_queries": [],
"validation_details": {
    "syntax_errors_found": 0,
    "queries_fixed": 0,
    "reasoning": "The provided Cypher query is syntactically correct and fully complies with all specified rules. It correctly matches the inlet node using its _id property, then creates all required equipment nodes with proper Neo4j label syntax, _label properties, and randomUUID() for _id assignment. Relationships are correctly defined using the allowed [:CONNECTED] type with appropriate properties. Variable scoping is valid, and the topology logically connects the inlet to the outlet through the specified process units. No RETURN clause is present, which is explicitly allowed. All constraints are satisfied without modification."
    }
}
\end{lstlisting}
\end{cypherbox}

\begin{schemebox}

\begin{lstlisting}[language=json]
{
"validation_status": true,
"validation_details": {
    "reasoning": "The scheme passes all structural checks: every node is connected, there are no orphan or duplicate nodes, all main equipment (pump, heat exchanger, separators) are flanked by ball valves, there are no cycles, no node has more than one incoming edge, and both separators have two outgoing streams. Logical consistency is satisfied as all streams carry the required thermophysical properties. No equipment node reports any violations, so physics validation is clear. The main outlet stream (edge 6530) delivers a mass flow of 136,813 kg/h, temperature 61.7 C, pressure 382,981 Pa, and composition (oil 0.9895, water 0.00999, gas 0.00052) which are all within the allowed 20% tolerance of the target outlet specifications (140,000 kg/h, 75 C, 450,000 Pa, oil 0.985, water 0.01, gas 0.005). Therefore the process meets the required performance criteria.",
    "outlet_flow_properties": {
        "mass_flow_kg_h": 136813.042,
        "temperature": 65.779,
        "pressure": 451088.342,
        "oil_frac": 0.98388,
        "water_frac": 0.00994,
        "gas_frac": 0.006182,
        "cp": 2120.053,
        "Q": 161.671,
        "rho": 846.243,
        "_fluid_type": "main"
        },
    "structural_issues": [],
    "logical_issues": [],
    "compliance_issues": [],
    "domain_rule_violations": [],
    "suggestions": []
    }
}
\end{lstlisting}
\end{schemebox}

\clearpage
\section{LLM prompt for hybrid method}
\label{app:llm prompt hybrid method}
This appendix provides an LLM prompt for the hybrid method which iteratively updates the PFD graph according to the validator's feedback.

\begin{lstlisting}[style=prompt-style, label=lst:prompt1]
You are a process flowsheet optimization expert.

EQUIPMENT DATABASE ACCESS

You have access to these tools:
- get_pumps
- get_heat_exchangers
- get_ball_valves
- get_remaining_equipment

MANDATORY TOOL POLICY:
- ALWAYS call ALL 4 tools before generating Cypher queries
- NEVER skip tools
- NEVER generate equipment not returned by tools
- If validation feedback contains: "Equipment used in scheme is insufficient" then call ALL tools again using adjusted constraints
- If the current equipment set cannot satisfy the targets, re-query tools or simplify the topology instead of fabricating equipment.

Tool selection logic:
- pumps -> pressure increase + flow rate
- heat exchangers -> temperature increase + pressure tolerance
- ball valves -> operating pressure
- remaining equipment -> separators and auxiliary units

Constraint strategy:
- Use realistic but not overly strict constraints, to increase the size of used equipment database - relax tools' parameters
- Prefer broader search ranges over exact matching
- Prefer feasible equipment over cheapest equipment
- Cost optimization is secondary to feasibility

Your task is to repair and optimize an existing process flow graph.

OBJECTIVES:
1. Minimize total equipment cost
2. Achieve target outlet flow properties (10% upward/downward deviations are allowed)
3. Resolve all physical violations
4. Preserve valid topology whenever possible - no orhan nodes, no cycles

RULES:
- Main equipment (pumps, separators, heat exchangers) must be flanked with Ball valves 
- Prefer local repairs over global redesign initially, but you are encouraged to perform moderate topology changes for optimization purposes (moving separators fron one place to another - in this case we can create flow spliting in various places)
- Do not modify equipment properties manually
- Only use equipment returned by tools
- Never create cycles, it's prohibited!
- Do not remove valid subgraphs
- Do not create orphan nodes!
- If pressure violation occurs:
  first reduce upstream pressure,
  then replace equipment if necessary
- If outlet flow mismatch occurs:
  prefer separators and tees before pumps
- You may delete unncessary equipment to minimize the cost (for instance, extra ball valves)
- Do not add pumps for gas streams
- If two identical valves are in series and one is redundant, delete the redundant valve instead of adding more equipment
- Adding more than one incoming edge to the nodes is prohibited!

Repair logic (FOLLOW ONLY THIS LOGIC!):
- target outlet pressure too low -> add an additional pump or replace the existing one with more powerful pump
- target outlet temperature too low -> add extra heat exchangers! 
- target outlet gas fraction too low -> do not add more separators; reduce gas separation only by removing one separator or choosing a less aggressive one
- target outlet water fraction too high -> add WaterSeparator stage
- target outlet mass flow rate is too high -> add pipe tee to split the flow rate evenly

SEPARATOR RULES

WaterSeparator:
- exactly 2 outgoing edges:
  - edge with stream_type="main"
  - edge with stream_type="water"
- water branch must end: WaterSeparator -> BallValve -> FlowOutPipeOffPageConnector

GasSeparator:
- exactly 2 outgoing edges:
  - edge with stream_type="main"
  - edge with stream_type="gas"
- gas branch must end: GasSeparator -> BallValve -> FlowOutPipeOffPageConnector

PIPETEE RULES:

PipeTee:
- exactly 2 outgoing edges with the same stream_type as in incoming edge
- one of the edges must end: PipeTee -> BallValve -> FlowOutPipeOffPageConnector

Do not reconnect water/gas branches to the main process.

You must return ONLY valid JSON.

GRAPH SCHEMA RULES

Node format:
When adding equipment returned by tools:

- Copy the ENTIRE equipment object exactly as returned.
- Do not omit fields.
- Do not simplify equipment definitions.
- Missing properties will cause solver failure.

{{
  "id": "5",
  "_label": "CentrifugalPump",
  **properties relevant for CentrifugalPump
}}

Edge format:
{{
  "id": 1,
  "from": "1",
  "to": "2",
  "stream_type": "main",
  "_type": "Pipe"
  "flow": flow properties in this edge
}}

Allowed patch keys only:
- replace_nodes: [{{node_id, new_equipment}}]
- add_nodes: [{{id, new_equipment}}]
- remove_nodes: [id, ...]
- add_edges: [{{from, to, stream_type}}]
- remove_edges: [id, ...]
Do not use new_node, _category, fluid_type, or any extra keys.

IMPORTANT:
- "stream_type" MUST be only "main", "water" or "gas"
- Use "_label" for equipment type
- Use "_type" for edge type
- Do not invent additional fields
- Do not generate flow properties
- Do not generate equipment properties manually

Allowed operations:
- replace_nodes
- add_nodes
- remove_nodes
- add_edges
- remove_edges

JSON FORMAT:

{{
  "reasoning": "...",
  "replace_nodes": [
    {{"id": "33" (id of the node that is being replaced), **set of new properties}},
    ...
  ],
  "add_nodes": [],
  "remove_nodes": [],
  "add_edges": [],
  "remove_edges": []
}}
\end{lstlisting}

\clearpage
\section{P\&ID Interaction Evaluation Prompts}
\label{app:pid prompts}
This appendix provides representative natural-language prompts used to evaluate the LLM-based interaction layer for the P\&ID graph model. Table~\ref{tab:pid-evaluation-prompts} lists ten representative rules covering different types of reasoning required for P\&ID analysis.

\begin{table}[h]
\centering
\caption{Representative P\&ID evaluation prompts used in the experiments.}
\label{tab:pid-evaluation-prompts}
\begin{tabular}{p{1.4cm}p{3.0cm}p{9.0cm}}
\toprule
Rule & Check type & Natural-language prompt \\
\midrule
Rule 4 &
Boundary, adjacency &
Check that the last element before the plant boundary is a shut-off valve or a control valve. \\

Rule 7 &
Equipment protection &
Check that each pump has shut-off valves on both the inlet and outlet sides. \\

Rule 14 &
Directed order, control loop &
On the reflux line, check that the flowmeter is located before the flow-control valve in the direction of flow and is linked to the control loop of this valve. \\

Rule 23 &
Local topology &
Check that a pressure indicator is connected to the discharge side of the pump. \\

Rule 48 &
Instrumentation placement &
Check that a temperature sensor is installed downstream of the heat exchanger on the main process line. \\

Rule 72 &
Drain connection &
Check that the low-point drain branch is connected through a valve and leads to the drain system. \\

Rule 104 &
Flare connection &
Check that the pressure-safety valve outlet is connected to the flare system. \\

Rule 136 &
Path, protective control &
Check that the centrifugal compressor has an anti-surge loop from the discharge side to the suction side. \\

Rule 214 &
Recycle branch &
Check that the recycle line branches from the main line and returns upstream without breaking the main flow path. \\

Rule 665 &
Forbidden bypass &
Check that the emergency shut-down valve used for fire isolation has no bypass path around it. \\
\bottomrule
\end{tabular}
\end{table}

\subsection{Examples of Generated Action Programs}

The following listings show two examples of action programs generated by the model. 

Listing~\ref{lst:pid-rule14-code} shows a directed-order check for a reflux line. The program verifies that the flowmeter is located before the flow-control valve and that it belongs to the same control loop.

\begin{lstlisting}[style=pythonblock, caption={Example generated code for Rule 14.}, label={lst:pid-rule14-code}]
violations = []

reflux_lines = sdk.find_edges_by_attribute("service", "reflux")

for line in reflux_lines:
    path = sdk.trace_path(
        start=line.source,
        direction="downstream",
        edge_filter={"service": "reflux"}
    )

    flowmeters = sdk.find_nodes_on_path(path, node_type="FlowMeter")
    control_valves = sdk.find_nodes_on_path(path, node_type="FlowControlValve")

    if not flowmeters or not control_valves:
        violations.append({
            "line": line.id,
            "reason": "Missing flowmeter or flow-control valve on reflux line"
        })
        continue

    flowmeter = flowmeters[0]
    control_valve = control_valves[0]

    if sdk.index_on_path(path, flowmeter.id) > sdk.index_on_path(path, control_valve.id):
        violations.append({
            "line": line.id,
            "reason": "Flowmeter is located after the flow-control valve"
        })

    linked_instruments = sdk.get_control_loop_members(control_valve.id)
    if flowmeter.id not in linked_instruments:
        violations.append({
            "line": line.id,
            "reason": "Flowmeter is not linked to the control loop of the valve"
        })

result = {
    "rule": "Rule 14",
    "passed": len(violations) == 0,
    "violations": violations
}
\end{lstlisting}

Listing~\ref{lst:pid-rule665-code} shows a path-based check for a forbidden bypass around a fire-isolation emergency shut-down valve.

\begin{lstlisting}[style=pythonblock, caption={Example generated code for Rule 665.}, label={lst:pid-rule665-code}]
violations = []

esv_nodes = sdk.find_nodes(
    node_type="EmergencyShutdownValve",
    attributes={"function": "fire_isolation"}
)

for esv in esv_nodes:
    upstream_nodes = sdk.get_neighbors(esv.id, direction="upstream")
    downstream_nodes = sdk.get_neighbors(esv.id, direction="downstream")

    if not upstream_nodes or not downstream_nodes:
        violations.append({
            "esv": esv.id,
            "reason": "Cannot determine upstream or downstream side of ESV"
        })
        continue

    upstream = upstream_nodes[0]
    downstream = downstream_nodes[0]

    bypass_paths = sdk.find_paths(
        start=upstream.id,
        target=downstream.id,
        direction="downstream",
        exclude_nodes=[esv.id],
        max_depth=12
    )

    if bypass_paths:
        violations.append({
            "esv": esv.id,
            "reason": "Bypass path exists around fire-isolation ESV",
            "paths": [sdk.path_to_ids(path) for path in bypass_paths]
        })

result = {
    "rule": "Rule 665",
    "passed": len(violations) == 0,
    "violations": violations
}
\end{lstlisting}

\section{Full Oil-Treatment PFD-to-P\&ID Enrichment}

\begin{figure*}[!t]
    \centering
    \includegraphics[width=\textwidth]
    {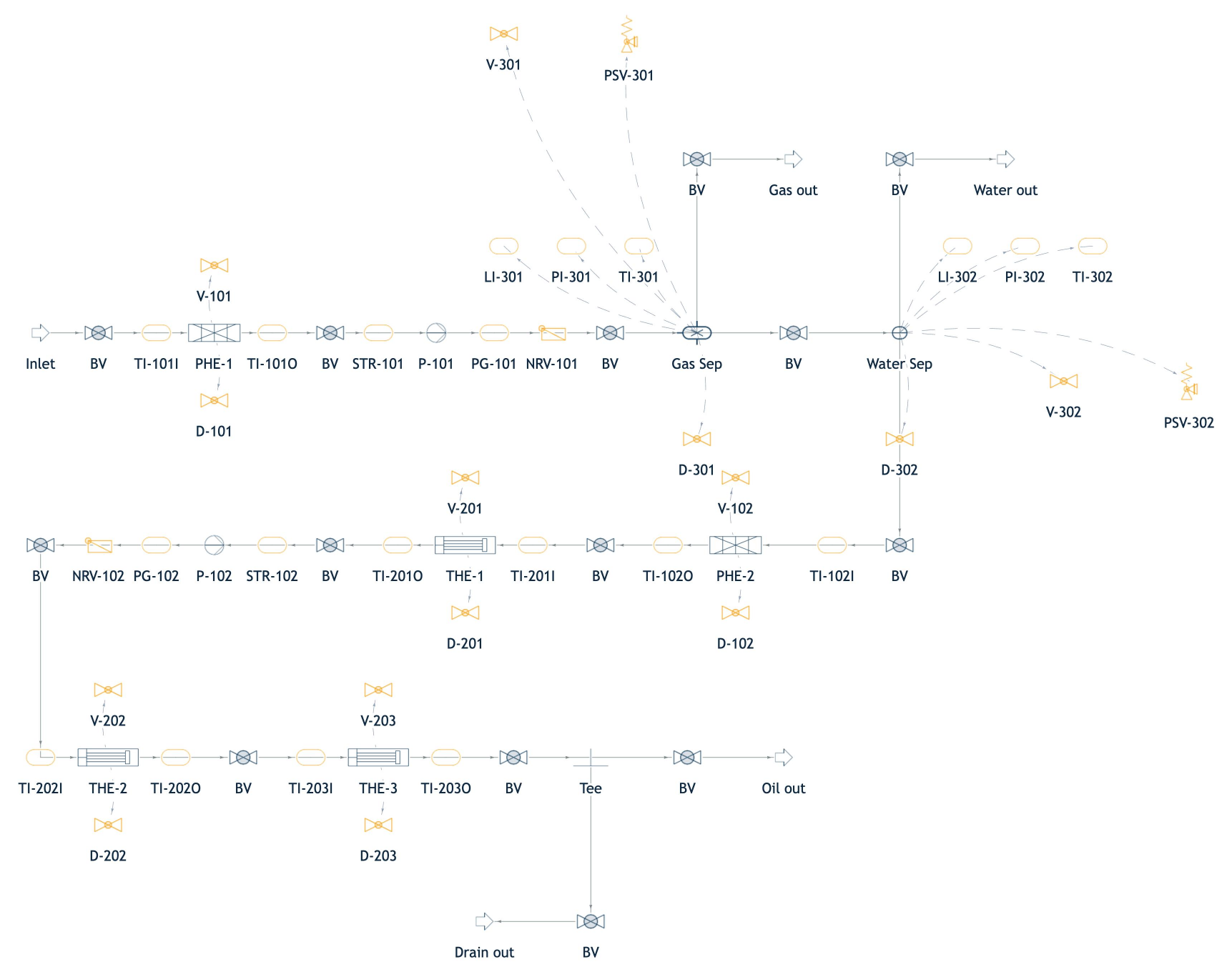}
    \caption{\textbf{Complete graph-level P\&ID enrichment of the oil-treatment
    PFD.}
    The figure shows the final graph after applying the selected
    source-backed enrichment rules to pumps, heat exchangers, separators, and
    outlet boundaries. Newly introduced P\&ID elements are highlighted in
    yellow. 
    }
    \label{fig:full-pid-enrichment}
\end{figure*}

Figure~\ref{fig:full-pid-enrichment} shows the complete enriched graph obtained for the oil-treatment case study.

As in the main text, the diagram is
shown as a graph representation rather than as a finalized plant layout, node
positions are arranged to make the added objects and their relations readable.

\end{document}